\pdfoutput=1

\documentclass[11pt]{article}

\usepackage[final]{acl}

\usepackage{times}
\usepackage{latexsym}

\usepackage[T1]{fontenc}

\usepackage[utf8]{inputenc}

\usepackage{microtype}

\usepackage{inconsolata}

\usepackage{graphicx}
\usepackage{booktabs}
\usepackage{multirow}
\usepackage{amsmath}
\usepackage{makecell}
\usepackage[table]{xcolor}
\usepackage{graphicx}
\usepackage{subcaption}
\usepackage{tabularx,array, xcolor}
\newcommand{\ph}[1]{\textcolor{teal!70!black}{\texttt{\{#1\}}}} 
\newcommand{\refrow}[1]{\textcolor{gray!90}{\textbf{#1}}}
\usepackage{arydshln}
\usepackage{amssymb}
\usepackage{pifont}
\newcommand{\cmark}{\checkmark}
\newcommand{\xmark}{\times}

\newcommand{\rotlalmmultirow}[2]{%
  \multirow{#1}{*}{\rotatebox[origin=c]{90}{\makecell[c]{#2}}}%
}

\title{To Trust or Not to Trust: Retrieval-Augmented Fact Checking in Speech}

\author{Debajyoti Mazumder$^1$, Mamta$^2$, Abhirama Subramanyam Penamakuri$^3$ \\
  $^1$IISER Bhopal, $^2$King's College London, $^3$MBZUAI\\
  {\small \texttt{debajyoti22@iiserb.ac.in, mamta20118@gmail.com, venkata.penamakuri@mbzuai.ac.ae}}\\
}

\begin{document}
\maketitle
\begin{abstract}
Online misinformation increasingly appears in spoken formats such as news clips, podcasts, interviews, political speeches, and social media videos, creating a need for fact-checking systems that can verify claims directly from speech. We introduce \textsc{VeriSpeak}, a probe benchmark for studying speech-based fact verification in Large Audio Language Models (LALMs). \textsc{VeriSpeak} contains 3,879 spoken claims spanning temporal, geographical, and relational facts, with balanced true and false labels. The benchmark is designed to examine whether factual verification ability transfers from text to speech, and whether retrieval-augmented LALMs can use textual evidence to correctly support or refute spoken claims. Our experiments reveal a consistent text-speech modality gap: LALMs that verify written claims reliably often fail on the same claims when spoken. Moreover, retrieval alone provides limited gains because models frequently conflate retrieved evidence with the spoken claim. In contrast, retrieval combined with explicit reasoning improves claim-evidence comparison, with a thinking-tuned LALM reaching 86.1\% accuracy. \textsc{VeriSpeak} highlights that effective speech misinformation detection requires not only speech understanding, but also grounded reasoning over retrieved evidence. 
The dataset is publicly available via Hugging Face at \url{https://huggingface.co/datasets/abhiram4572/VeriSpeak}.

\end{abstract}

\begin{figure}[t!]
   \centering
   \scriptsize
  \includegraphics[width=\columnwidth]{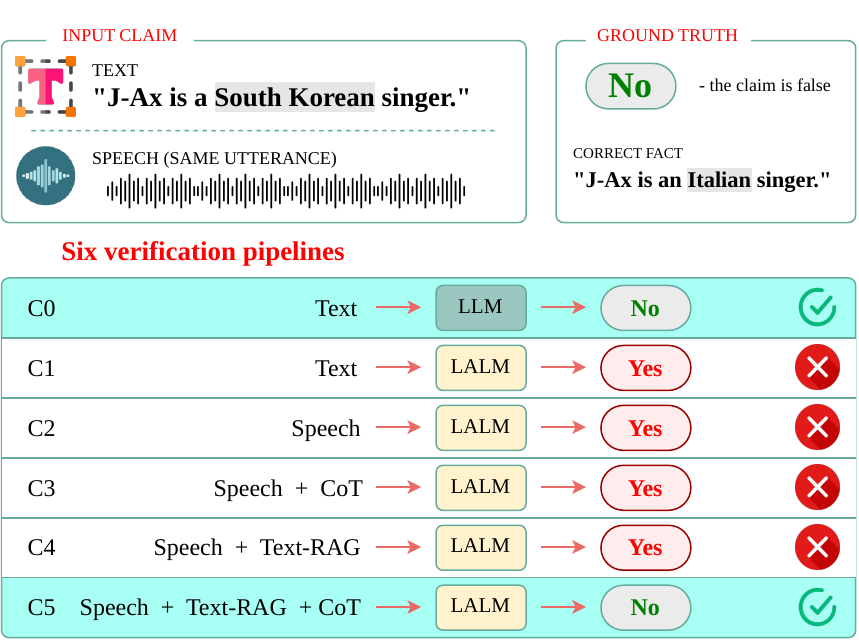}
\caption{\textbf{\textsc{VeriSpeak} probe pipelines for spoken fact verification.} Given the same false claim in text and speech, we compare six controlled settings that vary input modality, model interface, transcript-based Text-RAG, and chain-of-thought (CoT) reasoning. The example highlights three findings: speech input exposes a text-speech verification gap; retrieval alone can still conflate the spoken claim with retrieved evidence; and, among speech-input settings, Text-RAG with CoT preserves the claim-evidence distinction and recovers the correct verdict. Results shown use Audio-Flamingo-next-think~\cite{DBLP:afnextthink}.}
 \label{fig:teaser_fig}
\end{figure}

\section{Introduction}

Spoken media, from news and podcasts to speeches and social videos, often conveys factual claims that may be true, false, outdated, or misleading\footnote{\url{https://reutersinstitute.politics.ox.ac.uk/digital-news-report/2025}}. As Large Audio Language Models (LALMs) become increasingly capable of following instructions over speech and audio, they create new opportunities for spoken fact-checking, misinformation detection, and voice-based content moderation~\cite{DBLP:journals/corr/abs-2311-07919,DBLP:journals/corr/abs-2407-10759,DBLP:audioflamingo,ghosh2026audio,DBLP:journals/corr/abs-2503-01743}. These applications require more than speech recognition: models must identify the spoken claim and judge its factuality.

Fact verification is commonly formulated as evidence-based claim verification where a system retrieves evidence and predicts whether a claim is supported or refuted~\cite{DBLP:fever,DBLP:hover,DBLP:scifact,aly-etal-2021-fact,mamta-cocarascu-2025-facteval}. This retrieve-and-verify paradigm also underlies retrieval-augmented Large Language Models (LLM) methods for knowledge-intensive generation and verification~\cite{DBLP:ragoriginalpaper,DBLP:selfrag}. Multimodal fact-checking further studies cases where information is distributed across modalities, especially in image-text misinformation settings~\cite{mishrafactify,mocheg,akhtar-etal-2023-multimodal,khaliq-etal-2024-ragar}. 
In contrast, speech fact verification poses a distinct cross-modal setting: the claim is acoustic, while the supporting evidence is textual. The model must therefore preserve the spoken claim as the verification target while reasoning over retrieved text.

This raises a broader question: \emph{Can LALMs perform evidence-grounded fact verification when the claim is spoken?} Although many LALMs are built on pretrained LLM backbones, speech-based verification requires more than accessing factual knowledge from text-trained parameters: the model must identify the spoken claim, use retrieved textual evidence, and compare the two to classify the claim. Prior work on modality gaps shows that models can behave differently across modalities even when the semantic content is similar~\citep{DBLP:bridge,DBLP:seesaw,DBLP:blindfaith,xiang2025understanding}; here, such gaps may affect both spoken-claim understanding and evidence use.

To investigate this systematically, we introduce \textsc{VeriSpeak}, a benchmark of 3,879 spoken claims spanning temporal, geographical, and relational facts, with balanced correct and incorrect labels. Its design enables controlled comparisons across text-only verification, text-input LALMs, speech-input LALMs, retrieval-augmented verification, and reasoning-augmented verification.

Using \textsc{VeriSpeak}, we systematically evaluate five LALMs~\cite{DBLP:journals/corr/abs-2311-07919,DBLP:journals/corr/abs-2407-10759,ghosh2026audio,DBLP:afnextthink,DBLP:journals/corr/abs-2503-01743} across three model families under text, speech, retrieval, and reasoning settings. Specifically, we investigate: (i) \emph{whether text-based factual verification ability transfers to spoken claims}, (ii) \emph{where the text--speech modality gap arises and whether retrieval can compensate for it}, (iii) \emph{how transcript-based retrieval compares with direct audio-query retrieval}, and (iv) \emph{when explicit reasoning helps LALMs use retrieved evidence}.

Our findings include: (i) LALMs exhibit a large and consistent text-speech modality gap: factual verification ability observed on textual claims does not reliably transfer to spoken claims, even when the same LALM remains strong on text inputs. (ii) Retrieval alone only partially improves speech fact verification. Transcript-based RAG is generally more reliable than audio-query RAG, but standard LALMs often conflate the spoken claim with retrieved evidence, treating the evidence itself as the statement to verify. (iii) Reasoning helps mainly when evidence is available. Chain-of-thought prompting alone does not reliably improve standard speech-only verification, but RAG with explicit reasoning improves claim-evidence comparison. A thinking-tuned LALM benefits most strongly, reaching 86.1\% accuracy with retrieval and reasoning. Figure~\ref{fig:teaser_fig} summarizes these settings and illustrates the main claim-evidence conflation failure mode.

Overall, \textsc{VeriSpeak} provides a controlled testbed for analyzing LALM fact verification across text, speech, retrieval, and reasoning settings. Our results show that spoken fact verification requires not only speech recognition and evidence retrieval, but also robust claim-evidence separation.

\section{Related Work}

\noindent \textbf{Fact Verification.} Fact verification has long relied on retrieval-augmented verification: a system first retrieves relevant evidence and then uses it to verify a claim. Benchmarks such as FEVER~\cite{DBLP:fever}, HoVer~\cite{DBLP:hover}, SciFact~\cite{DBLP:scifact}, and FEVEROUS~\cite{aly-etal-2021-fact} instantiate this broad retrieve-and-verify paradigm across diverse inputs. While retrieval modules vary from sparse retrieval to dense neural retrieval, performance depends on two factors: whether the system retrieves useful evidence, and whether the verifier can correctly leverage the retrieved evidence for verification. Modern RAG-style systems generalize this retrieve-and-condition paradigm to LLM-based generation, verification, revision, and critique~\cite{DBLP:ragoriginalpaper,DBLP:selfrag,gao-etal-2023-rarr}.

Recent multimodal fact verification extends evidence-based verification beyond text, mainly to visual--text misinformation~\cite{mocheg,mishrafactify,chakraborty-etal-2023-factify3m,akhtar-etal-2023-multimodal,khaliq-etal-2024-ragar,kangur-etal-2025-multireflect}. These works show that verification becomes harder when claims and evidence span modalities. Speech fact verification poses a distinct challenge: the claim is spoken, while the retrieved evidence is typically textual. We study this setting through a controlled benchmark, asking whether Large Audio Language Models (LALMs) can leverage retrieved textual evidence for spoken claims as reliably as text-only LLMs do for written claims.

\noindent \textbf{Large Audio Language Models.} LALMs couple pretrained language models with speech or audio encoders, enabling models to follow instructions over spoken and non-speech audio inputs. Early systems such as Pengi~\cite{deshmukh2023pengi}, LTU-AS~\cite{gong_ltuas}, and SpeechGPT~\cite{zhang2023speechgpt} showed that LLMs can be adapted to audio-conditioned interaction. Recent models, including AudioPaLM~\cite{DBLP:journals/corr/abs-2306-12925}, SALMONN~\cite{tang2024salmonn}, WavLLM~\cite{hhu2024wavllm}, GAMA~\cite{ghosh-etal-2024-gama}, MERaLiON-AudioLLM~\cite{DBLP:journals/corr/abs-2412-09818}, Qwen-Audio~\cite{DBLP:journals/corr/abs-2311-07919}, Qwen2-Audio~\cite{DBLP:journals/corr/abs-2407-10759}, Audio-Flamingo~\cite{DBLP:audioflamingo}, Audio-Flamingo-3~\cite{ghosh2026audio}, and Phi-4-Multimodal~\cite{DBLP:journals/corr/abs-2503-01743}, further improve speech understanding, audio reasoning, and instruction following. However, these capabilities do not directly imply reliable factual verification. Speech fact verification requires a model to recover the claim from audio, preserve it as the target of verification, and reason over external textual evidence without conflating the two sources. Our work evaluates whether current LALMs can meet this requirement.

\noindent\textbf{Multimodal Modality Gap.} Prior work has shown that multimodal models can suffer from modality gaps and modality imbalance, where one modality is not used as reliably as another~\cite{DBLP:bridge,DBLP:seesaw,DBLP:mbpo}. Similar issues arise in Large Audio/Speech Language Models: the same linguistic content can lead to different behavior depending on whether it is presented as text or speech~\cite{xiang2025understanding}. Existing work mainly studies this problem through representation alignment, token-level matching, or cross-modal calibration~\cite{issam2025dtw,DBLP:journals/corr/abs-2510-13632}.

We study a complementary question: whether speech-capable models can perform evidence-grounded verification when the claim is spoken and the evidence is textual. In this setting, the model must maintain a clear separation between the spoken claim and retrieved evidence, then reason over both sources to decide veracity.

\section{VeriSpeak: A Probe-Benchmark for Speech Fact Verification}
\label{sec:verispeak_dataset}
 
To probe factual reasoning capabilities of LALMs in a controlled setting, we construct \textsc{VeriSpeak} through a systematic multi-stage pipeline that maintains strict control over data quality, label distribution, and category balance.
 
\noindent \textbf{Knowledge Base and Enrichment.} We build on the knowledge base
of~\cite{Shah_Mishra_Yadati_Talukdar_2019}, which provides short free-text
celebrity biographies keyed by entity IDs, with the entity name appearing at the
start of each record. Using this name, we scrape
Wikipedia\footnote{\url{https://www.wikipedia.org/}} to enrich each biography with
structured metadata, resolving birth country and gender and disambiguating umbrella
entries such as the United Kingdom into their constituent nations (England,
Scotland, Wales, Northern Ireland). Targeted audit passes then repair null or
misresolved fields, including disambiguation over multiple query variants and
normalization of historical country labels (e.g., ``Kingdom of Great Britain''
$\to$ ``United Kingdom''). 
 
\noindent \textbf{Fact Category Flagging.} We define three categories that require
qualitatively different knowledge types, letting us diagnose whether the
text-speech gap is uniform or category-specific: \textit{temporal} facts (years and
dates), \textit{geographical} facts (locations, nationalities), and
\textit{relational} facts (employment, kinship, organizational affiliations). Year
mentions are detected by regular expressions over four-digit patterns; location
mentions via spaCy named-entity recognition~\cite{honnibal2020spacy} over geographic spans; and relation
mentions by matching marital, parental, and sibling keywords (\textit{married},
\textit{spouse}, \textit{wife/husband}, \textit{son/daughter of},
\textit{father/mother of}, \textit{brother/sister of}) at the sentence level.
 
\noindent \textbf{Atomic Fact Extraction.} To ensure model errors reflect reasoning
failures rather than claim ambiguity, each probe item must be a single, unambiguous
claim. For each flagged celebrity, we prompt Llama-3.2-3B-Instruct~\cite{grattafiori2024llama} to produce
atomic, pronoun-free sentences grounded in the bio, anchored on the relevant
temporal, geographical, or relational signal. Outputs containing fewer than four words or unresolved pronouns are
discarded during post-processing; full filtering criteria are provided
in Appendix~\ref{app:atomic-prompts}.
 
\noindent \textbf{Controlled Negative Generation.} A balanced binary probe requires
plausible incorrect counterparts; implausible negatives would let models exploit
surface cues rather than genuine reasoning. We generate negatives automatically
using deterministic, category-specific perturbations. For \textit{temporal} facts,
the year is shifted by a non-zero offset in $[-10, +10]$, clamped to a plausible
four-digit range. For \textit{geographical} facts, named-entity recognition
identifies country, city, and nationality spans, each swapped with a same-type
candidate drawn from typed pools over the full knowledge base, excluding the
celebrity's own country. For \textit{relational} facts, we apply two strategies
with equal probability: (1) a \textit{relation-word swap} using a hand-built map
(\textit{married to} $\to$ \textit{divorced from}, \textit{son of} $\to$
\textit{father of}, etc.); and (2) an \textit{entity swap}, replacing person or organization spans with a random same-type knowledge-base entity, never the celebrity themselves. Celebrities with no valid negative are discarded, keeping balanced labels.

\noindent \textbf{Speech Synthesis.} To isolate the effect of the audio modality from confounds such as speaker variation, all claims are synthesized using the Coqui TTS engine with the \texttt{tacotron2-DDC} vocoder~\cite{DBLP:tachotron2}, producing natural-sounding audio with consistent pronunciation and intonation. A single-speaker model is used deliberately to prevent models from exploiting speaker identity as a spurious verification cue. 
 
\noindent \textbf{Probe Composition.} The resulting \textsc{VeriSpeak} probe set comprises 3,879 items spanning three fact categories (year: 1,451; location: 2,226; relation: 202), each with a balanced split of correct and incorrect claims. Each item includes: (1) the audio file, (2) an automatic transcript generated by Wav2Vec 2.0~\cite{DBLP:wav2vec}, and (3) the ground-truth veracity label. To reflect realistic deployment conditions, we use automatic transcripts 
despite the resulting transcription errors. Although this increases probe difficulty, it ensures observed failures genuinely reflect multimodal reasoning limitations rather than artifacts of perfect input.

\begin{figure}[h]
    \centering
    \includegraphics[width=\linewidth]{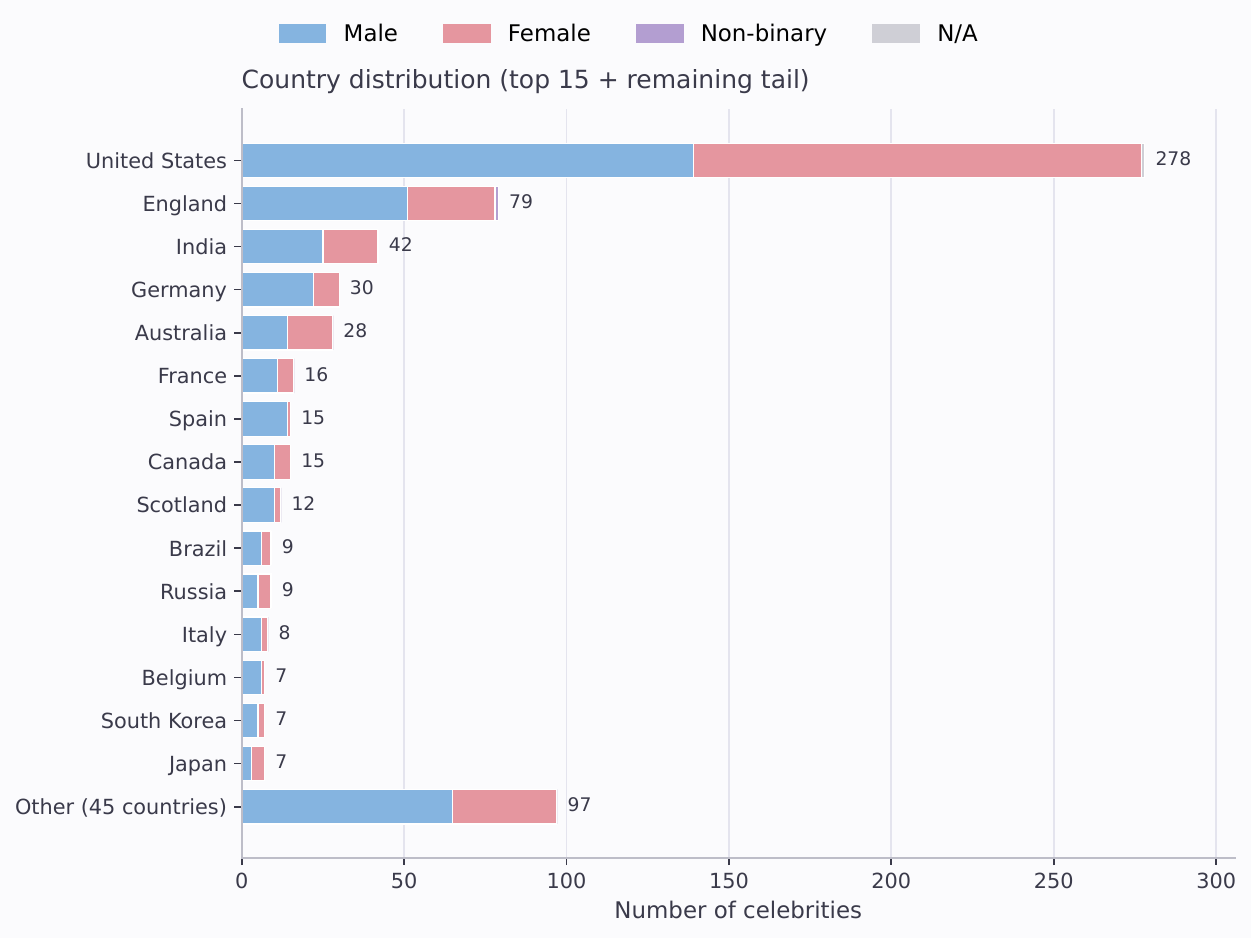}
    \caption{Demographic breakdown of \textsc{VeriSpeak}'s 659 subject entities.}
    \label{fig:demography_gender}
\end{figure}

\noindent \textbf{Dataset Demographics.}
Figure~\ref{fig:demography_gender} presents the country distribution of the 659 subject entities (celebrities) whose facts comprise \textsc{VeriSpeak}. The subjects span 60 countries, with the largest representation from the United States (278) and England (79), followed by a diverse set of countries across multiple regions. This distribution reflects the composition of the underlying knowledge base. Gender representation comprises 59.5\% male (392), 40.2\% female (265), 0.2\% non-binary (1), and 0.2\% N/A (1).

\section{Experimental Setup}




\label{sec:experimental-setup}


\paragraph{Models.}
As shown in Table~\ref{tab:models}, we evaluate speech-capable Large Audio Language Models (LALMs) from three model families: Qwen-7B~\cite{DBLP:qwen}, Qwen2.5-7B~\cite{DBLP:qwem25}, and Phi-4-mini-instruct~\cite{DBLP:journals/corr/abs-2503-01743}. The Qwen family includes Qwen-Audio-Chat~\cite{DBLP:journals/corr/abs-2311-07919} and Qwen2-Audio-7B~\cite{DBLP:journals/corr/abs-2407-10759}; Qwen2.5 includes Audio-Flamingo-3~\cite{ghosh2026audio}; and Phi includes Phi-4-multimodal~\cite{DBLP:journals/corr/abs-2503-01743}. For each LALM, we use its associated text-only LLM backbone as a reference. This pairing lets us test whether factual verification ability observed in the text backbone remains accessible when the same claims are given as speech. We evaluate each LALM with both text and speech inputs to separate text-mode performance from speech-input degradation. We also evaluate Audio-Flamingo-next-think~\cite{DBLP:afnextthink} to examine whether explicit reasoning improves speech-based fact verification.

\begin{table}[t]
\centering
\small
\resizebox{\columnwidth}{!}{%
\begin{tabular}{lll}
\toprule
\textbf{Family} & \textbf{LLM-backbone} & \textbf{Speech Model} \\
\midrule
\multirow{2}{*}{Qwen} 
& \multirow{2}{*}{Qwen-7B} 
& Qwen-Audio-Chat \\
& & Qwen2-Audio-7B-Instruct \\

\midrule

\multirow{2}{*}{Qwen2.5} 
& \multirow{2}{*}{Qwen2.5-7B-Instruct} 
& Audio-Flamingo-3 \\
& & Audio-Flamingo-next-think \\

\midrule

Phi 
& Phi-4-mini-instruct 
& Phi-4-multimodal-instruct \\
\bottomrule
\end{tabular}%
}
\caption{Model families evaluated in our experiments. 
}
\label{tab:models}
\end{table}














\paragraph{Evaluation conditions.}
We evaluate each model under controlled conditions that vary the input modality, the model interface, the availability of retrieved evidence, and the use of explicit reasoning. 

\textsc{Text-LLM} ($\mathbf{T}_{\mathrm{LLM}}$; C0) evaluates the text-only LLM on the written claim. This is the reference condition for factual verification in the backbone's native text modality.

\textsc{Text-LALM} ($\mathbf{T}_{\mathrm{LALM}}$; C1) evaluates the LALM on the same written claim. This condition tests whether the LALM preserves the text-side verification ability of its associated LLM backbone.

\textsc{Speech-LALM} ($\mathbf{S}_{\mathrm{LALM}}$; C2) evaluates the LALM on the spoken claim without retrieved evidence. This is the main speech fact verification.

\textsc{Speech-LALM-CoT} ($\mathbf{S}_{\mathrm{LALM}}[\textsc{CoT}]$; C3) evaluates the LALM on the spoken claim with chain-of-thought prompting, but without retrieved evidence. This condition tests whether explicit reasoning helps the model verify the spoken claim before adding external context.



\begin{figure}[t!]
   \centering
   \scriptsize
  \includegraphics[width=\columnwidth]{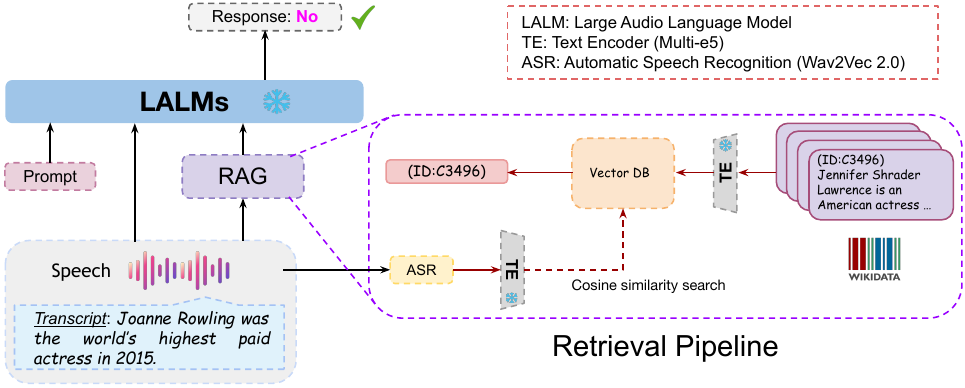}
\caption{\textbf{\textsc{Transcript-RAG} pipeline.} ASR transcribed spoken claim is used to retrieve textual evidence, which is leveraged by LALM to verify the spoken claim.}
 \label{fig:transcript_rag}
\end{figure}

\paragraph{Retrieval variants.} 
\textsc{Transcript-RAG} ($\mathbf{S}^{\mathbf{r}_{\text{text}}}_{\text{LALM}}$; C4) evaluates the LALM on the spoken claim together with textual evidence retrieved using the claim transcript. Prior work has shown the benefit of external knowledge for multimodal reasoning across visual~\cite{gatti-etal-2022-cofar,penamakuri-mishra-2024-visual} and audio~\cite{audiopedia} settings; here, we study its role in spoken fact verification. Figure~\ref{fig:transcript_rag} illustrates this pipeline: retrieval is performed using the ASR (Automatic Speech Recognition) transcript, while the spoken claim remains the input to be verified. Specifically, the audio claim is first transcribed using wav2vec 2.0~\cite{DBLP:wav2vec}, and the resulting transcript is used to retrieve textual evidence from the knowledge base. We use multi-e5~\cite{DBLP:multie5} as the default retriever. To test sensitivity to retriever choice, we additionally evaluate e5-large-v2~\cite{DBLP:e5} and msmarco-MiniLM-L12-v3~\cite{reimers-gurevych-2019-sentence}.

\textsc{Transcript-RAG-CoT} ($\mathbf{S}_{\mathrm{LALM}}^{\mathbf{r}_{\mathrm{text}}}[\textsc{CoT}]$; C5) evaluates the same transcript-based RAG condition with chain-of-thought prompting. This condition tests whether explicit reasoning helps the model leverage retrieved evidence while keeping the spoken claim as the object of verification. 

We also evaluate an audio-query retrieval variant, \textsc{AudioQuery-RAG} ($\mathbf{S}_{\mathrm{LALM}}^{\mathbf{r}_{\mathrm{audio}}}$), where the audio claim itself is used as the retrieval query. We use a CLAP audio-text retriever~\cite{DBLP:conf/icassp/ElizaldeDIW23} that maps audio queries and textual evidence into a shared embedding space.

For RQ6, we additionally evaluate \textsc{Text-RAG-LLM} ($\mathbf{T}_{\mathrm{LLM}}^{\mathbf{r}_{\mathrm{text}}}$), where the text-only LLM receives the written claim together with multi-e5 retrieved evidence. This serves as an upper bound. Prompts for all these evaluation conditions are included in Appendix (Table~\ref{tab:prompts}).


\paragraph{Metrics and comparisons.}
We report accuracy for each factual category and average accuracy across \textit{Year}, \textit{Location}, and \textit{Relation}. Let $A(\cdot)$ denote average accuracy. For each model pair, we report two comparison metrics:
\begin{equation}
\Delta
=
A(\mathbf{T}_{\mathrm{LLM}})
-
A(\text{setting}).
\label{eq:delta}
\end{equation}
\begin{equation}
{Lift}
=
A(\text{setting})
-
A(\mathbf{S}_{\mathrm{LALM}}).
\label{eqn:delta}
\end{equation}

Here, $\mathbf{T}_{\mathrm{LLM}}$ is the paired text-only LLM baseline and
$\mathbf{S}_{\mathrm{LALM}}$ is the paired speech-only LALM.
Positive $\Delta$ indicates degradation relative to the text-only baseline,
while negative $\Delta$ indicates improvement over it. Positive
${Lift}$ indicates improvement over the speech-only LALM baseline. In CoT settings, outputs that omit a parseable verdict in the required \texttt{<answer>} field are scored as incorrect.


\section{Research Questions}
\label{sec:research-questions}

Using the setup described previously, we now study the following research questions.

\noindent \textbf{RQ1. Does factual verification transfer reliably from text to speech?}

\noindent \textbf{Test.}
We compare the text-only reference setting $\mathbf{T}_{\mathrm{LLM}}$ with the speech-only LALM setting $\mathbf{S}_{\mathrm{LALM}}$ across model families and factual categories. In Table~\ref{tab:main_results}, this corresponds to comparing the rows without retrieval and without CoT prompting.

\noindent \textbf{Finding.}
No. LALMs show a large and consistent modality gap. Qwen-7B achieves 75.3\% accuracy in the text-only setting, while Qwen-Audio-Chat and Qwen2-Audio-7B reach only 50.4\% and 49.2\% in the speech-only setting, corresponding to $\Delta$ values of 24.9 and 26.1 points. Audio-Flamingo-3 drops from 71.8\% to 58.5\%, yielding a 13.3 point gap, and Phi-4-multimodal drops from 68.0\% to 49.2\%, yielding an 18.8 point gap.

The gap appears across all model families, indicating that factual verification behavior available in text does not reliably transfer when the same claims are presented as speech. This suggests a speech-side knowledge accessibility problem: the relevant factual knowledge may be available to the model family, but it is not consistently activated through the speech interface.

\noindent \textbf{RQ2. Where does the text-speech modality gap arise?}

\noindent \textbf{Test.}
We decompose the drop from the text-only reference condition C0 to the speech-only condition C2 into two local components. The first captures the text-mode difference between the LLM backbone (C0) and the LALM (C1): $\Delta_{\mathrm{text}} = A(\mathbf{T}_{\mathrm{LLM}}) - A(\mathbf{T}_{\mathrm{LALM}}).$
The second captures the additional drop caused by replacing text input with speech input within the same LALM (C2):
$\Delta_{\mathrm{speech}}=A(\mathbf{T}_{\mathrm{LALM}}) - A(\mathbf{S}_{\mathrm{LALM}}).$
Thus, the table's $\Delta$ value for $\mathbf{S}_{\mathrm{LALM}}$ is the sum of these two components: $\Delta=\Delta_{\mathrm{text}} + \Delta_{\mathrm{speech}}.$

\noindent \textbf{Finding.}
Text-mode differences vary across models, but speech-input degradation is consistent. In Table~\ref{tab:main_results}, Qwen-Audio-Chat and Audio-Flamingo-3 have positive $\Delta_{\mathrm{text}}$ values of 9.7 and 6.8 points, respectively, indicating lower text-mode performance than their reference LLMs. This suggests that multimodal adaptation may degrade text-side factual verification ability. In contrast, Qwen2-Audio-7B and Phi-4-multimodal have negative $\Delta_{\mathrm{text}}$ values of $-1.1$ and $-4.4$ points, meaning that they match or slightly outperform their text-only backbones when evaluated on text. Thus, text-side degradation is not universal across LALMs.

However, all models drop when the same LALM receives speech instead of text. The $\Delta_{\mathrm{speech}}$ values are 15.2 points for Qwen-Audio-Chat, 27.2 points for Qwen2-Audio-7B, 6.5 points for Audio-Flamingo-3, and 23.2 points for Phi-4-multimodal. This shows that the modality gap is not simply a uniform loss of text-mode factual ability after multimodal adaptation. A major source of degradation is the speech interface itself: the LALM often fails to elicit the same factual verification behavior from spoken claims that it can produce from written claims.

This motivates retrieval as a natural intervention. If parametric factual knowledge is not reliably accessed from speech, then retrieved textual evidence may help compensate it. 

\begin{table}[t!]
\centering
\scriptsize
\setlength{\tabcolsep}{2pt}
\renewcommand{\arraystretch}{1.25}
\resizebox{\columnwidth}{!}
{%
\begin{tabular}{@{}cclccrrrrrr@{}}
\toprule
\textbf{LALM} 
& \textbf{Cond.}
& \textbf{Setting} 
& \textbf{Retr.}
& \textbf{CoT}
& \textbf{Year} 
& \textbf{Loc.} 
& \textbf{Rel.} 
& \textbf{Avg.}
& \textbf{$\Delta$ ($\downarrow$)}
& \textbf{$Lift$ ($\uparrow$)}\\
\midrule

\multicolumn{11}{c}{\cellcolor[gray]{0.9}\textbf{Qwen-7B as LLM backbone}} \\
\midrule

\rotlalmmultirow{6}{Qwen-Audio-Chat}
& \refrow{C0}
& \refrow{$\mathbf{T}_{\mathrm{LLM}}$} 
& \refrow{$\xmark$} & \refrow{$\xmark$}
& \refrow{60.4} & \refrow{87.5} & \refrow{78.2} & \refrow{75.3} & \refrow{--} & {-} \\
& C1
& $\mathbf{T}_{\mathrm{LALM}}$ 
& $\xmark$ & $\xmark$
& \cellcolor{red!12}53.4 & \cellcolor{red!22}73.1 & \cellcolor{red!13}70.3 & 65.6 & 9.7 & - \\
& C2
& $\mathbf{S}_{\mathrm{LALM}}$ 
& $\xmark$ & $\xmark$
& \cellcolor{red!15}\underline{51.5} & \cellcolor{red!40}50.1 & \cellcolor{red!35}\underline{49.5} & 50.4 & 24.9 & - \\
& C3
& $\mathbf{S}_{\mathrm{LALM}}[\textsc{CoT}]$
& $\xmark$ & $\cmark$
& \cellcolor{red!33}40.4 & \cellcolor{red!43}46.6 & \cellcolor{red!39}47.5 & 44.9 & 30.4 & $-5.5$ \\
& C4
& $\mathbf{S}_{\mathrm{LALM}}^{\mathbf{r}_{\mathrm{text}}}$
& $\cmark$ & $\xmark$
& \cellcolor{red!16}50.8 & \cellcolor{red!38}\textbf{54.2} & \cellcolor{red!34}\textbf{50.5} & \underline{51.8} & \underline{23.5} & \underline{1.4} \\
& C5
& $\mathbf{S}_{\mathrm{LALM}}^{\mathbf{r}_{\mathrm{text}}}[\textsc{CoT}]$
& $\cmark$ & $\cmark$
& \cellcolor{red!9}\textbf{54.9} & \cellcolor{red!41}\underline{52.8} & \cellcolor{red!35}\textbf{50.5} & \textbf{52.8} & \textbf{22.5} & \textbf{2.4} \\

\midrule

\rotlalmmultirow{6}{Qwen2-Audio-7B}
& \refrow{C0}
& \refrow{$\mathbf{T}_{\mathrm{LLM}}$} 
& \refrow{$\xmark$} & \refrow{$\xmark$}
& \refrow{60.4} & \refrow{87.5} & \refrow{78.2} & \refrow{75.3} & \refrow{--} & - \\
& C1
& $\mathbf{T}_{\mathrm{LALM}}$ 
& $\xmark$ & $\xmark$
& \cellcolor{red!3}59.3 & \cellcolor{green!8}88.3 & \cellcolor{green!12}81.7 & 76.4 & $-1.1$ & - \\
& C2
& $\mathbf{S}_{\mathrm{LALM}}$ 
& $\xmark$ & $\xmark$
& \cellcolor{red!20}47.7 & \cellcolor{red!40}\underline{50.0} & \cellcolor{red!35}50.0 & 49.2 & 26.1 & - \\
& C3
& $\mathbf{S}_{\mathrm{LALM}}[\textsc{CoT}]$
& $\xmark$ & $\cmark$
& \cellcolor{red!28}43.5 & \cellcolor{red!49}44.7 & \cellcolor{red!47}45.1 & 44.4 & 30.9 & $-4.8$ \\
& C4
& $\mathbf{S}_{\mathrm{LALM}}^{\mathbf{r}_{\mathrm{text}}}$
& $\cmark$ & $\xmark$
& \cellcolor{red!14}\underline{52.2} & \cellcolor{red!40}\underline{50.0} & \cellcolor{red!34}\underline{50.5} & \underline{50.9} & \underline{24.4} & \underline{1.7} \\
& C5
& $\mathbf{S}_{\mathrm{LALM}}^{\mathbf{r}_{\mathrm{text}}}[\textsc{CoT}]$
& $\cmark$ & $\cmark$
& \cellcolor{red!3}\textbf{58.4} & \cellcolor{red!33}\textbf{58.6} & \cellcolor{red!31}\textbf{54.0} & \textbf{57.0} & \textbf{18.3} & \textbf{7.8} \\

\midrule

\multicolumn{11}{c}{\cellcolor[gray]{0.9}\textbf{Qwen2.5-7B as LLM backbone}} \\
\midrule

\rotlalmmultirow{6}{Audio-Flamingo-3}
& \refrow{C0}
& \refrow{$\mathbf{T}_{\mathrm{LLM}}$} 
& \refrow{$\xmark$} & \refrow{$\xmark$}
& \refrow{63.3} & \refrow{78.4} & \refrow{73.8} & \refrow{71.8} & \refrow{--} & - \\
& C1
& $\mathbf{T}_{\mathrm{LALM}}$ 
& $\xmark$ & $\xmark$
& \cellcolor{red!10}57.9 & \cellcolor{red!9}73.4 & \cellcolor{red!16}63.9 & 65.0 & 6.8 & - \\
& C2
& $\mathbf{S}_{\mathrm{LALM}}$ 
& $\xmark$ & $\xmark$
& \cellcolor{red!15}54.2 & \cellcolor{red!25}61.5 & \cellcolor{red!22}\underline{59.9} & 58.5 & 13.3 & - \\
& C3
& $\mathbf{S}_{\mathrm{LALM}}[\textsc{CoT}]$
& $\xmark$ & $\cmark$
& \cellcolor{red!17}52.6 & \cellcolor{red!20}62.9 & \cellcolor{red!25}55.5 & 57.0 & 14.8 & $-1.5$ \\
& C4
& $\mathbf{S}_{\mathrm{LALM}}^{\mathbf{r}_{\mathrm{text}}}$
& $\cmark$ & $\xmark$
& \cellcolor{red!6}\underline{60.0} & \cellcolor{red!23}\underline{63.0} & \cellcolor{red!25}56.9 & \underline{60.0} & \underline{11.8} & \underline{1.5} \\
& C5
& $\mathbf{S}_{\mathrm{LALM}}^{\mathbf{r}_{\mathrm{text}}}[\textsc{CoT}]$
& $\cmark$ & $\cmark$
& \cellcolor{red!5}\textbf{60.4} & \cellcolor{red!13}\textbf{71.3} & \cellcolor{red!12}\textbf{64.9} & \textbf{65.5} & \textbf{6.3} & \textbf{7.0} \\

\midrule

\multicolumn{11}{c}{\cellcolor[gray]{0.9}\textbf{Phi-4-mini-instruct as LLM backbone}} \\
\midrule

\rotlalmmultirow{6}{Phi-4-multimodal} 
& \refrow{C0}
& \refrow{$\mathbf{T}_{\mathrm{LLM}}$} 
& \refrow{$\xmark$} & \refrow{$\xmark$}
& \refrow{56.3} & \refrow{74.8} & \refrow{72.8} & \refrow{68.0} & \refrow{--} & - \\
& C1
& $\mathbf{T}_{\mathrm{LALM}}$ 
& $\xmark$ & $\xmark$
& \cellcolor{green!14}60.5 & \cellcolor{green!18}81.3 & \cellcolor{green!10}75.3 & 72.4 & $-4.4$ & - \\
& C2
& $\mathbf{S}_{\mathrm{LALM}}$ 
& $\xmark$ & $\xmark$
& \cellcolor{red!14}51.2 & \cellcolor{red!35}55.3 & \cellcolor{red!32}53.5 & 53.3 & 14.7 & - \\
& C3
& $\mathbf{S}_{\mathrm{LALM}}[\textsc{CoT}]$
& $\xmark$ & $\cmark$
& \cellcolor{red!40}16.3 & \cellcolor{red!60}20.4 & \cellcolor{red!60}13.4 & 16.7 & 51.3 & $-36.6$ \\
& C4
& $\mathbf{S}_{\mathrm{LALM}}^{\mathbf{r}_{\mathrm{text}}}$
& $\cmark$ & $\xmark$
& \cellcolor{red!3}\underline{55.1} & \cellcolor{red!29}\underline{54.5} & \cellcolor{red!30}\underline{52.0} & \underline{53.9} & \underline{14.1} & \underline{0.6} \\
& C5
& $\mathbf{S}_{\mathrm{LALM}}^{\mathbf{r}_{\mathrm{text}}}[\textsc{CoT}]$
& $\cmark$ & $\cmark$
& \cellcolor{green!34}\textbf{65.9} & \cellcolor{red!18}\textbf{61.7} & \cellcolor{red!25}\textbf{54.5} & \textbf{60.7} & \textbf{7.3} & \textbf{7.4} \\

\bottomrule
\end{tabular}%
}
\caption{
Main results on \textsc{VeriSpeak}. $\Delta$ is the drop from $\mathbf{T}_{\mathrm{LLM}}$; lower is better. Lift is the gain over the paired speech-only baseline $\mathbf{S}_{\mathrm{LALM}}$; higher is better. Cell colors reflect $\Delta$, green for gains and red for drops. Bold/underline denote the best/second-best speech-input scores for each LALM.
}
\label{tab:main_results}
\end{table}

\begin{table}[t]
\centering
\scriptsize
\setlength{\tabcolsep}{2pt}
\renewcommand{\arraystretch}{1.3}
\resizebox{\columnwidth}{!}
{%
\begin{tabular}{@{}llrrrrrr@{}}
\toprule
\textbf{LALM} 
& \textbf{Setting} 
& \textbf{Year} 
& \textbf{Location} 
& \textbf{Relation} 
& \textbf{Avg.}
& \textbf{$\Delta$} ($\downarrow$)
& \textbf{$Lift$} ($\uparrow$) \\
\midrule

\multicolumn{8}{c}{\cellcolor[gray]{0.9}\textbf{Qwen-7B as LLM backbone}} \\
\midrule

\multirow{3}{*}{\makecell{Qwen-\\Audio-Chat}} 
& $\mathbf{S}_{\mathrm{LALM}}$ 
& 51.5 & 50.1 & 49.5 & 50.4 & 24.9 & - \\
& $\mathbf{S}_{\mathrm{LALM}}^{\mathbf{r}_{\mathrm{text}}}$ 
& 50.8 & 54.2 & 50.5 & \textbf{51.8} & \textbf{23.5} & \textbf{+1.4} \\
& $\mathbf{S}_{\mathrm{LALM}}^{\mathbf{r}_{\mathrm{audio}}}$ 
& 48.5 & 50.0 & 50.0 & 49.5 & 25.8 & $-0.9$ \\

\midrule

\multirow{3}{*}{\makecell{Qwen2-\\Audio-7B}} 
& $\mathbf{S}_{\mathrm{LALM}}$ 
& 47.7 & \textbf{50.0} & 50.0 & 49.2 & 26.1 & - \\
& $\mathbf{S}_{\mathrm{LALM}}^{\mathbf{r}_{\mathrm{text}}}$ 
& \textbf{52.2} & \textbf{50.0} & \textbf{50.5} & \textbf{50.9} & \textbf{24.4} & \textbf{+1.7} \\
& $\mathbf{S}_{\mathrm{LALM}}^{\mathbf{r}_{\mathrm{audio}}}$ 
& 52.3 & 50.0 & \textbf{50.0} & 50.8 & 24.5 & +1.6 \\

\midrule

\multicolumn{8}{c}{\cellcolor[gray]{0.9}\textbf{Qwen2.5-7B as LLM backbone}} \\
\midrule

\multirow{3}{*}{\makecell{Audio-\\Flamingo-3}} 
& $\mathbf{S}_{\mathrm{LALM}}$ 
& 54.2 & 61.5 & \textbf{59.9} & 58.5 & 13.3 & - \\
& $\mathbf{S}_{\mathrm{LALM}}^{\mathbf{r}_{\mathrm{text}}}$ 
& \textbf{60.0} & \textbf{63.0} & 56.9 & \textbf{60.0} & \textbf{11.8} & \textbf{+1.5} \\
& $\mathbf{S}_{\mathrm{LALM}}^{\mathbf{r}_{\mathrm{audio}}}$ 
& 52.0 & 56.3 & 52.5 & 53.6 & 18.2 & $-4.9$ \\

\midrule

\multicolumn{8}{c}{\cellcolor[gray]{0.9}\textbf{Phi-4-mini-instruct as LLM backbone}} \\
\midrule

\multirow{3}{*}{\makecell{Phi-4-\\multimodal}} 
& $\mathbf{S}_{\mathrm{LALM}}$ 
& 51.2 & 55.3 & 53.5 & 53.3 & 14.7 & - \\
& $\mathbf{S}_{\mathrm{LALM}}^{\mathbf{r}_{\mathrm{text}}}$ 
& \textbf{55.1} & \textbf{54.5} & \textbf{52.0} & \textbf{53.9} & \textbf{14.1} & \textbf{+0.6} \\
& $\mathbf{S}_{\mathrm{LALM}}^{\mathbf{r}_{\mathrm{audio}}}$ 
& 52.8 & 52.3 & 51.5 & 52.2 & 15.8 & -1.1 \\

\bottomrule
\end{tabular}%
}
\caption{
\textsc{Transcript-RAG} vs. \textsc{AudioQuery-RAG} 
}
\label{tab:rag_query_comparison}
\end{table}

\noindent \textbf{RQ3. Can retrieval compensate for the text-speech modality gap?}

\noindent \textbf{Test.}
Motivated by the speech-input degradation observed in RQ2, we compare the speech-only condition $\mathbf{S}_{\mathrm{LALM}}$ with two retrieval-augmented conditions: \textsc{Transcript-RAG} $\mathbf{S}_{\mathrm{LALM}}^{\mathbf{r}_{\mathrm{text}}}$ and \textsc{AudioQuery-RAG} $\mathbf{S}_{\mathrm{LALM}}^{\mathbf{r}_{\mathrm{audio}}}$. Table~\ref{tab:rag_query_comparison} reports category-level accuracy, average accuracy, \textit{$Lift$}, and $\Delta$ for these settings. 
If retrieval compensates for the text-speech modality gap, we expect high positive $Lift$ and a substantially reduced $\Delta$.

\noindent \textbf{Finding.} 
Retrieval helps, but only partially. As shown in Table~\ref{tab:rag_query_comparison}, \textsc{Transcript-RAG} with multi-e5 improves all standard LALMs over their speech-only baselines: Qwen-Audio-Chat improves from 50.4\% to 51.8\%, Qwen2-Audio-7B from 49.2\% to 50.9\%, Audio-Flamingo-3 from 58.5\% to 60.0\%, and Phi-4-multimodal from 49.2\% to 53.9\%. However, these gains are modest: $Lift$ ranges from 1.4 to 4.7 points and averages only 2.3 points across standard LALMs.

The remaining gaps to the text-only baseline are still large. After \textsc{Transcript-RAG}, $\Delta$ remains 23.5 points for Qwen-Audio-Chat, 24.4 for Qwen2-Audio-7B, 11.8 for Audio-Flamingo-3, and 14.1 for Phi-4-multimodal. Thus, transcript-based retrieval improves speech fact verification, but does not close the text-speech modality gap.

\noindent \textbf{Retriever analysis.}
Table~\ref{tab:retriever_ablation} further compares retrieval choices. Among transcript-based retrievers, multi-e5 is the most reliable overall, although msmarco performs best for Qwen-Audio-Chat. In contrast, \textsc{AudioQuery-RAG} with CLAP is weaker for standard LALMs: its average $Lift$ is $-0.3$ points, compared with $+2.3$ for \textsc{transcript-RAG} with multi-e5 in Table~\ref{tab:rag_query_comparison}. This is consistent with CLAP's low Recall@1 of approximately 0.3\%, meaning that the top-ranked passage often fails to provide the relevant evidence.

Overall, retrieval alone is insufficient for standard LALMs. The bottleneck is not only evidence availability or retrieval quality, but also the model's ability to reason over the spoken claim and the retrieved context. We therefore next ask whether explicit reasoning improves speech fact verification before studying reasoning in the retrieval-augmented setting.

\noindent \textbf{Diagnostic Analysis.}
To understand why standard RAG yields only limited gains, we test whether retrieved evidence changes the model's perceived object of verification. We sample 50 Qwen2-Audio-7B-Instruct \textsc{Transcript-RAG} examples (from the factually incorrect sample pool) and ask the model, given the spoken claim and retrieved evidence, to extract the claim to be fact-checked. These extracted claims are manually labeled as matching the spoken claim, the retrieved evidence, or other/ambiguous.

\noindent \textbf{Finding.}
RAG often shifts verification away from the spoken claim. Only \textbf{31\%} of extracted claims match the original claim, while \textbf{65\%} match the retrieved evidence, with the remaining 4\% under other/ambiguous. Thus, the model more often treats the retrieved passage as the claim rather than as evidence and returns `Yes'. This explains the limited RAG gains in RQ3: even when relevant evidence is retrieved, the LALM fails to preserve the spoken claim as the verification target and compare it against the evidence. When it instead verifies the retrieved text itself, retrieval cannot help.

\begin{table}[t!]
\centering
\small
\setlength{\tabcolsep}{4pt}
\renewcommand{\arraystretch}{1.08}
\resizebox{\columnwidth}{!}{%
\begin{tabular}{@{}lrrrrr@{}}
\toprule
\multirow{2}{*}{\textbf{LALM}} 
& \multirow{2}{*}{$\mathbf{S}_{\mathrm{LALM}}$}
& \multicolumn{3}{c}{$\mathbf{S}_{\mathrm{LALM}}^{\mathbf{r}_{\mathrm{text}}}$}
& $\mathbf{S}_{\mathrm{LALM}}^{\mathbf{r}_{\mathrm{audio}}}$ \\
\cmidrule(lr){3-5}
\cmidrule(l){6-6}
& 
& \textbf{multi-e5} 
& \textbf{e5-v2} 
& \textbf{msmarco} 
& \textbf{CLAP} \\
\midrule
Recall@1 (\%) & -- & 85.6 & 78.2 & 49.8 & 0.3 \\

\midrule

Qwen-Audio-Chat 
& 50.4 & 51.8 & 53.0 & \textbf{53.8} & 49.5 \\

Qwen2-Audio-7B 
& 49.2 & \textbf{50.9} & 50.7 & 50.6 & 50.8 \\

Phi-4-multimodal 
& 49.2 & \textbf{53.9} & 53.4 & 53.8 & 52.2 \\

Audio-Flamingo-3 
& 58.5 & \textbf{60.0} & 59.6 & 57.6 & 53.6 \\

AF-next-think
& 66.4 & \textbf{81.4} & 80.5 & 74.1 & 60.2 \\

\bottomrule
\end{tabular}%
}
\caption{
Retriever ablation results on \textsc{VeriSpeak}.}
\label{tab:retriever_ablation}
\end{table}

\noindent \textbf{RQ4. Does chain-of-thought prompting improve speech-only fact verification?}

\noindent \textbf{Test.}
We compare C2, the speech-only condition $\mathbf{S}_{\mathrm{LALM}}$, with C3, its chain-of-thought variant $\mathbf{S}_{\mathrm{LALM}}[\textsc{CoT}]$. This tests whether explicit reasoning helps the LALM verify the spoken claim without any retrieved evidence.

\noindent \textbf{Finding.}
No. CoT alone does not improve speech-only verification for standard LALMs. As shown in Table~\ref{tab:main_results}, C3 performs worse than C2 for all models: Qwen-Audio-Chat drops from 50.4\% to 44.9\%, Qwen2-Audio-7B from 49.2\% to 44.4\%, Audio-Flamingo-3 from 58.5\% to 57.0\%, and Phi-4-multimodal from 49.2\% to 16.7\% (Phi-4-multimodal's low C3 result is caused by format failure: only 29.8\% of outputs contain a parseable verdict. Full diagnostics are in Appendix~\ref{app:output-diagnostics}). Correspondingly, the average $\Delta$ increases from 20.8 to 31.9 points across models, indicating that CoT moves the models farther from their text-only baselines. 

Thus, simply asking a standard LALM to reason step by step is not enough to recover factual verification ability from speech. Without external evidence, CoT can even destabilize the verification decision. This motivates the next question: whether reasoning becomes useful when the model is given retrieved evidence to reason over.

\noindent \textbf{RQ5. Does explicit reasoning help LALMs leverage retrieved evidence for verification?}

\noindent \textbf{Test.}
We compare C4, \textsc{Transcript-RAG} $\mathbf{S}_{\mathrm{LALM}}^{\mathbf{r}_{\mathrm{text}}}$, with C5, \textsc{Transcript-RAG-CoT} $\mathbf{S}_{\mathrm{LALM}}^{\mathbf{r}_{\mathrm{text}}}[\textsc{CoT}]$. This tests whether CoT helps the model use retrieved textual evidence to verify the spoken claim.

\noindent \textbf{Finding.}
Yes. Unlike CoT alone, CoT with retrieval consistently improves performance. In Table~\ref{tab:main_results}, C5 improves over C4 for every standard LALM: Qwen-Audio-Chat improves from 51.8\% to 52.8\%, Qwen2-Audio-7B from 50.9\% to 57.0\%, Audio-Flamingo-3 from 60.0\% to 65.5\%, and Phi-4-multimodal from 53.9\% to 60.7\%. On average, adding CoT to \textsc{Transcript-RAG} improves accuracy by 4.9 points.

The same trend appears in the main $\Delta$ metric. C5 reduces the remaining gap to the text-only baseline from 23.5 to 22.5 points for Qwen-Audio-Chat, from 24.4 to 18.3 points for Qwen2-Audio-7B, from 11.8 to 6.3 points for Audio-Flamingo-3, and from 14.1 to 7.3 points for Phi-4-multimodal. Averaged across models, CoT reduces $\Delta$ by 4.9 points when retrieval is present.

This contrast between RQ4 and RQ5 is central: CoT alone does not help speech-only verification, but CoT helps once retrieved evidence is available. This suggests that explicit reasoning is most useful for claim--evidence comparison, not for recovering factual knowledge from speech alone. In other words, reasoning helps the LALM leverage retrieved evidence for the correct purpose: verifying the spoken claim rather than treating the retrieved text as the claim itself.

\begin{table}[t]
\centering
\small
\setlength{\tabcolsep}{4pt}
\renewcommand{\arraystretch}{1.10}
\resizebox{\columnwidth}{!}{%
\begin{tabular}{lrrrrrrr}
\toprule
\textbf{LALM}
& \textbf{C2}
& \textbf{C3}
& \textbf{C4}
& \textbf{C5}
& \textbf{$Lift$} ($\uparrow$)
& $\Delta$ ($\downarrow$)
& $\Delta_{\mathrm{UB}}$ ($\downarrow$) \\
\midrule

AF-3
& \cellcolor{red!20}58.5
& \cellcolor{red!23}57.0
& \cellcolor{red!18}60.0
& \cellcolor{red!9}65.5
& +7.0 & 6.3 & 26.9 \\

AF-next-think
& \cellcolor{red!7}\textbf{66.4}
& \cellcolor{green!0}\textbf{71.8}
& \cellcolor{green!16}\textbf{81.4}
& \cellcolor{green!24}\textbf{86.1}
& \textbf{+19.7} & {$\textbf{-14.3}$} & \textbf{6.3} \\

\midrule

\rowcolor{gray!10}
\multicolumn{8}{@{}c@{}}{
$\mathbf{T}_{\mathrm{LLM}} = 71.8$,
\quad
$\mathbf{T}_{\mathrm{LLM}}^{\mathbf{r}_{\mathrm{text}}} = 92.4$
} \\

\bottomrule
\end{tabular}%
}
\caption{
Comparison between a standard LALM (AF-3) and a thinking-tuned LALM (AF-next-think).
}
\label{tab:reasoning_rag}
\end{table}

\noindent \textbf{RQ6. Do thinking-tuned LALMs make retrieval more effective for speech fact verification?}

\noindent \textbf{Test.}
We compare a standard LALM, Audio-Flamingo-3, with a thinking-tuned LALM, Audio-Flamingo-next-think, under the same four speech settings: C2 ($\mathbf{S}_{\mathrm{LALM}}$), C3 ($\mathbf{S}_{\mathrm{LALM}}[\textsc{CoT}]$), C4 ($\mathbf{S}_{\mathrm{LALM}}^{\mathbf{r}_{\mathrm{text}}}$), and C5 ($\mathbf{S}_{\mathrm{LALM}}^{\mathbf{r}_{\mathrm{text}}}[\textsc{CoT}]$). This comparison separates inference-time CoT prompting from model-level thinking ability. We report $Lift$ from C2 to C5, the remaining gap $\Delta$ to the Qwen2.5-7B text-only baseline, and the upper-bound gap: $\Delta_{\mathrm{UB}}=A(\mathbf{T}_{\mathrm{LLM}}^{\mathbf{r}_{\mathrm{text}}})-A(\mathrm{C5}).$

\noindent \textbf{Finding.}
Yes. Table~\ref{tab:reasoning_rag} shows that the thinking-tuned model benefits from reasoning and retrieval much more strongly than the standard model. For standard Audio-Flamingo-3, vanilla CoT does not help: C3 drops slightly below C2, from 58.5\% to 57.0\%. Retrieval alone improves performance to 60.0\%, and RAG+CoT reaches 65.5\%, giving a $Lift$ of 7.0 points from C2 to C5.

In contrast, Audio-Flamingo-next-think shows gains at every step. Vanilla CoT already improves speech-only verification from 66.4\% to 71.8\%, unlike the pattern observed for standard LALMs in RQ4. \textsc{Transcript-RAG} further improves performance to 81.4\%, and RAG+CoT reaches 86.1\%, giving a much larger $Lift$ of 19.7 points from C2 to C5. Further, on $\Delta$, Audio-Flamingo-next-think with C5 exceeds the text-only baseline by 14.3 points, giving a negative $\Delta$. This means that, once retrieval and CoT are combined with a thinking-tuned LALM, speech-based verification can outperform the non-retrieval text-only baseline. However, the model remains 6.3 points below the text-only RAG upper bound $\mathbf{T}_{\mathrm{LLM}}^{\mathbf{r}_{\mathrm{text}}}$, showing that there is still a measurable gap between speech-based RAG and fully text-based RAG.

Overall, thinking-specific instruction tuning changes how the model benefits from both CoT and retrieval. For standard LALMs, CoT alone is unreliable and retrieval gives modest gains. For the thinking-tuned LALM, CoT improves speech-only verification, RAG provides a large additional gain, and RAG+CoT gives the strongest result. This suggests that reasoning-trained LALMs are better able to preserve the spoken claim, compare it against retrieved textual evidence, and produce a reliable verification decision.



\begin{table}[t] 
\centering 
\small 
\resizebox{\columnwidth}{!}{ 
\begin{tabular}{lcccc} 
\toprule 
\textbf{LALM} & \textbf{Year} & \textbf{Location} & \textbf{Relation} & \textbf{Overall} \\ 
\midrule 
Qwen-Audio-Chat & 0.25 & 0.24 & 0.21 & 0.23 \\ 
Qwen2-Audio-7B-Instruct & 0.16 & 0.19 & 0.21 & 0.18 \\ 
Phi-4-multimodal & 0.16 & 0.20 & 0.27 & 0.21 \\ 
Audio-Flamingo-3 & {0.11} & {0.13} & {0.14} & {0.13} \\ 
Audio-Flamingo-next-think & {0.11} & 0.16 & 0.15 & 0.14 \\ 
\bottomrule 
\end{tabular}} 
\caption{Word error rate (WER; lower is better) of LALM-generated transcriptions across claim categories.} 
\label{tab:transcription-wer} 
\end{table}

\section{Additional Analysis}

\paragraph{WER Analysis.} To distinguish speech-perception errors from downstream verification errors, we compute the word error rate (WER) between each LALM's generated transcription and the original claim text. For year claims, we normalize equivalent numerical expressions (e.g., ``2008'' and ``two thousand eight'') before scoring. As shown in Table~\ref{tab:transcription-wer}, overall WER ranges from 0.13 to 0.23, indicating model-specific variation in recovering the spoken content. 

As a human intelligibility reference, we recruited two native English speakers to transcribe 50 randomly sampled clips. Their transcriptions yielded an avg. WER of 0.105, with annotator-level WERs of 0.15 and 0.06, suggesting that the synthesized speech is generally intelligible to human listeners. To characterize the remaining model errors, we further analyze the outputs of AF-3. Approximately 74\% of its WER edit operations involve subject names or other proper nouns, often reflecting spelling or phonetic variants such as ``Richie/Ritchie,'' ``Kris/Chris,'' and ``Bachelet/Bachellet.'' In contrast, non-name tokens have an error rate of approximately 3\%. Thus, the main challenge is entity recognition rather than general audio intelligibility. This distinction is particularly important for fact verification, where names and locations are often label-critical.

\paragraph{Transcription-based verification contrast.} To estimate the downstream effect of LALMs speech-recognition errors, we add an auxiliary diagnostic condition, C0$'$. In C0$'$, each LALM-generated transcription is provided as text input to the same paired text LLM evaluated in C0. Thus, C0 and C0$'$ differ only in whether the LLM receives the original claim or the model-generated transcription. We define $\Delta_{\mathrm{recognition}} = A(\text{C0}) - A(\text{C0}')$ as an approximate estimate of the verification degradation attributable to speech misrecognition.

The results show a measurable degradation, with $\Delta_{\mathrm{recognition}}$ ranging from 5.6 to 11.6 points across models. This trend is broadly consistent with the WER analysis, where transcription quality varies across LALMs. However, WER alone does not fully determine verification performance, as errors affecting fact-critical information (e.g., entities, years, locations, or relations) can have a disproportionate impact on the final prediction.



\begin{table}[t] 
\centering 
\footnotesize
\resizebox{\columnwidth}{!}{ 
\begin{tabular}{lccc} 
\toprule 
\textbf{LALM transcript source} & C0 & C0$'$ & ${\Delta_{\mathrm{recognition}}}$ 
\\ 
\midrule 
Qwen-Audio-Chat & 75.4 & 63.8 & 11.6 \\ 
Qwen2-Audio-7B & 75.4 & 65.2 & 10.2 \\ 
Phi-4-multimodal & 68.0 & 60.0 & 8.0 \\ 
Audio-Flamingo-3 & 71.8 & 63.4 & 8.4 \\ 
Audio-Flamingo-next-think & 71.8 & 66.2 & {5.6} 
\\ 
\bottomrule 
\end{tabular} 
}
\caption{Accuracy (\%) under the original-text condition (C0) and the LALM-transcription contrast (C0$'$).} 
\label{tab:transcription-contrast} 
\end{table}

\section{Conclusion and Future Work}

We introduced \textsc{VeriSpeak}, a controlled benchmark for evidence-grounded fact verification in speech, containing 3,879 spoken claims across temporal, geographical, and relational facts. Our experiments show that factual verification ability in text does not reliably transfer to speech: LALMs exhibit a consistent text--speech modality gap, even when their text-side performance remains strong. We further find that retrieval alone only partially addresses this gap, since standard LALMs often conflate retrieved textual evidence with the spoken claim itself. In contrast, retrieval combined with explicit reasoning improves claim--evidence comparison, with a thinking-tuned LALM reaching 86.1\% accuracy under transcript-based RAG with CoT. These results suggest that robust speech fact-checking requires not only speech recognition and evidence retrieval, but also mechanisms for preserving the spoken claim, grounding it in external evidence, and reasoning across modalities. 

\noindent \textbf{Future Work.}
Looking ahead, a broader goal for speech fact-checking is to build LALMs with little or no degradation relative to strong text-only LLM baselines. Ideally, LALMs should not rely on retrieval merely to compensate for weak speech-side factual access; when the required knowledge is available, spoken claims should be verified as reliably as written ones. For open-world or time-sensitive claims, retrieval will remain necessary, but models must better preserve the spoken claim as the verification target, treat retrieved text only as evidence, and explicitly compare the two. Although the thinking-tuned model narrows the gap, reaching 86.1\% with transcript-based RAG and CoT, it still trails the fully text-based RAG upper bound, leaving speech--text verification parity as an important research direction. In our future work, we aim to extend \textsc{VeriSpeak} to multilingual claims, diverse accents and dialects, natural speech, broader fact domains, and richer evidence sources.


\section*{Limitations}

\textsc{VeriSpeak} is designed as a controlled probe benchmark, and its limitations largely follow from this design choice. The benchmark focuses on temporal, geographical, and relational facts from celebrity biographies, enabling controlled analysis across fact types but leaving broader claim types such as scientific, medical, numerical, causal, and multi-hop claims for future work. The spoken claims are synthesized with a single-speaker TTS system to isolate the effect of input modality, which means the benchmark does not capture spontaneous speech, background noise, diverse accents, dialects, or multilingual speech. Retrieval is also performed over a fixed evidence collection to support controlled claim--evidence comparisons, rather than over noisy, conflicting, or time-sensitive open-world sources. More broadly, speech-based fact verification remains a largely open direction in speech NLP, and these limitations point to several important avenues for future research.

\section*{Ethical Considerations}
\textsc{VeriSpeak} is derived from the publicly available KVQA knowledge base, which contains public-figure information sourced from Wikidata, and does not include private or user-provided personal data. The benchmark contains synthetic false claims solely for controlled evaluation; each claim is paired with a veracity label and clearly identified as benchmark-generated in the dataset documentation and metadata. \textsc{VeriSpeak} is intended for evaluating speech-based fact-verification systems and should not be treated as a source of factual claims about individuals.

\section*{Acknowledgments}
Abhirama thanks his postdoc advisors, Yova Kementchedjhieva and Thamar Solorio at MBZUAI for supporting this work.  

\bibliography{custom}

\clearpage
\appendix

\section{Additional Analysis}
\label{sec:addn_analysis}
Tables~\ref{tab:score-combined-lift} and~\ref{tab:score-combined-delta} provide a more fine-grained view of how retrieval and reasoning affect different factual and demographic slices. The $Lift$ table (Table.~\ref{tab:score-combined-lift}) measures how much \textsc{Transcript-RAG+CoT} improves over the corresponding speech-only LALM, while the $\Delta$ table (Table~\ref{tab:score-combined-delta}) measures the remaining gap to the paired text-only LLM baseline.

\begin{table*}[t!]
\centering
\scriptsize
\setlength{\tabcolsep}{4pt}
\resizebox{\textwidth}{!}{
\begin{tabular}{llccccccccc}
\toprule
 &  & \multicolumn{3}{c}{Category} &  \multicolumn{2}{c}{Gender} &  \multicolumn{4}{c}{Country} \\
\cmidrule(lr){3-5} \cmidrule(lr){6-7} \cmidrule(lr){8-11}
Family & Model & Year & Loc. & Rel. & Male & Female & USA & England & India & Germany \\
\midrule

$\mathbf{T}_\mathrm{LLM}$ 
& Qwen-7B 
& 60.37 & 87.47 & 78.22 & 78.33 & 74.52 & 74.07 & 79.14 & 82.96 & 72.73 \\
& Qwen2.5-7B-Instruct 
& 63.34 & 78.39 & 73.76 & 73.64 & 70.74 & 70.03 & 77.51 & 78.03 & 68.18 \\
& Phi-4-mini-instruct 
& 56.31 & 74.80 & 72.77 & 69.83 & 64.42 & 67.01 & 73.21 & 77.58 & 62.27 \\

\midrule

$\mathbf{S}_\mathrm{LALM}$ 
& Qwen-Audio-Chat 
& 51.48 & 50.13 & 49.50 & 50.39 & 50.96 & 51.28 & 49.90 & 50.67 & 47.27 \\
& Qwen2-Audio-7B-Instruct 
& 47.69 & 50.00 & 50.00 & 49.23 & 48.97 & 48.65 & 49.69 & 49.78 & 50.00 \\
& Audio-Flamingo-3 
& 54.24 & 61.46 & 59.90 & 59.18 & 57.90 & 59.18 & 58.08 & 58.74 & 54.09 \\
& AF-Next-Think 
& 63.06 & 69.72 & 66.34 & 67.22 & 66.90 & 64.96 & 72.60 & 68.16 & 65.00 \\

\midrule

$\mathbf{S}_\mathrm{LALM}^{\mathbf{r}_\mathrm{text}}[\textsc{CoT}]$ 
& Qwen-Audio-Chat 
& $54.93_{\scriptscriptstyle +3.45}$ 
& $52.83_{\scriptscriptstyle +2.70}$ 
& $50.50_{\scriptscriptstyle +1.00}$ 
& $54.16_{\scriptscriptstyle +3.77}$ 
& $52.40_{\scriptscriptstyle +1.44}$ 
& $53.59_{\scriptscriptstyle +2.31}$ 
& $51.94_{\scriptscriptstyle +2.04}$ 
& $53.81_{\scriptscriptstyle +3.14}$ 
& $51.82_{\scriptscriptstyle +4.55}$ \\

& Qwen2-Audio-7B-Instruct 
& $58.37_{\scriptscriptstyle +10.68}$ 
& $58.63_{\scriptscriptstyle +8.63}$ 
& $53.96_{\scriptscriptstyle +3.96}$ 
& $59.30_{\scriptscriptstyle +10.07}$ 
& $56.66_{\scriptscriptstyle +7.69}$ 
& $60.46_{\scriptscriptstyle +11.81}$ 
& $56.24_{\scriptscriptstyle +6.55}$ 
& $52.02_{\scriptscriptstyle +2.24}$ 
& $52.27_{\scriptscriptstyle +2.27}$ \\

& Audio-Flamingo-3 
& $60.37_{\scriptscriptstyle +6.13}$ 
& $71.25_{\scriptscriptstyle +9.79}$ 
& $64.85_{\scriptscriptstyle +4.95}$ 
& $67.34_{\scriptscriptstyle +8.16}$ 
& $66.00_{\scriptscriptstyle +8.10}$ 
& $63.86_{\scriptscriptstyle +4.68}$ 
& $68.30_{\scriptscriptstyle +10.22}$ 
& $70.85_{\scriptscriptstyle +12.11}$ 
& $59.55_{\scriptscriptstyle +5.46}$ \\

& AF-Next-Think 
& $88.56_{\scriptscriptstyle +25.50}$ 
& $86.66_{\scriptscriptstyle +16.94}$ 
& $83.17_{\scriptscriptstyle +16.83}$ 
& $87.48_{\scriptscriptstyle +20.26}$ 
& $86.68_{\scriptscriptstyle +19.78}$ 
& $85.82_{\scriptscriptstyle +20.86}$ 
& $89.57_{\scriptscriptstyle +16.97}$ 
& $89.24_{\scriptscriptstyle +21.08}$ 
& $85.00_{\scriptscriptstyle +20.00}$ \\

\midrule
\multicolumn{2}{c}{\textbf{Avg. $Lift$ ($\uparrow$)}} 
& $\mathbf{+11.44}$ 
& $\mathbf{+9.51}$ 
& $\mathbf{+6.68}$ 
& $\mathbf{+10.57}$ 
& $\mathbf{+9.25}$ 
& $\mathbf{+9.92}$ 
& $\mathbf{+8.95}$ 
& $\mathbf{+9.64}$ 
& $\mathbf{+8.07}$ \\

\bottomrule
\end{tabular}
}
\caption{Verification accuracy (\%) across categories, gender, and country. For each \textsc{CoT} result, the subscript reports the absolute $Lift$ over the corresponding speech-only LALM baseline. Top-4 countries are shown.}
\label{tab:score-combined-lift}
\end{table*}

\begin{table*}[t!]
\centering
\scriptsize
\setlength{\tabcolsep}{4pt}
\resizebox{\textwidth}{!}{
\begin{tabular}{llccccccccc}
\toprule
 &  & \multicolumn{3}{c}{Category} &  \multicolumn{2}{c}{Gender} &  \multicolumn{4}{c}{Country} \\
\cmidrule(lr){3-5} \cmidrule(lr){6-7} \cmidrule(lr){8-11}
Family & Model & Year & Loc. & Rel. & Male & Female & USA & England & India & Germany \\
\midrule

$\mathbf{T}_\mathrm{LLM}$ 
& Qwen-7B 
& 60.37 & 87.47 & 78.22 & 78.33 & 74.52 & 74.07 & 79.14 & 82.96 & 72.73 \\
& Qwen2.5-7B-Instruct 
& 63.34 & 78.39 & 73.76 & 73.64 & 70.74 & 70.03 & 77.51 & 78.03 & 68.18 \\
& Phi-4-mini-instruct 
& 56.31 & 74.80 & 72.77 & 69.83 & 64.42 & 67.01 & 73.21 & 77.58 & 62.27 \\

\midrule

$\mathbf{S}_\mathrm{LALM}$ 
& Qwen-Audio-Chat 
& 51.48 & 50.13 & 49.50 & 50.39 & 50.96 & 51.28 & 49.90 & 50.67 & 47.27 \\
& Qwen2-Audio-7B-Instruct 
& 47.69 & 50.00 & 50.00 & 49.23 & 48.97 & 48.65 & 49.69 & 49.78 & 50.00 \\
& Audio-Flamingo-3 
& 54.24 & 61.46 & 59.90 & 59.18 & 57.90 & 59.18 & 58.08 & 58.74 & 54.09 \\
& AF-Next-Think 
& 63.06 & 69.72 & 66.34 & 67.22 & 66.90 & 64.96 & 72.60 & 68.16 & 65.00 \\

\midrule

$\mathbf{S}_\mathrm{LALM}^{\mathbf{r}_\mathrm{text}}[\textsc{CoT}]$ 
& Qwen-Audio-Chat 
& $54.93_{\scriptscriptstyle +5.44}$ 
& $52.83_{\scriptscriptstyle +34.64}$ 
& $50.50_{\scriptscriptstyle +27.72}$ 
& $54.16_{\scriptscriptstyle +24.17}$ 
& $52.40_{\scriptscriptstyle +22.12}$ 
& $53.59_{\scriptscriptstyle +20.48}$ 
& $51.94_{\scriptscriptstyle +27.20}$ 
& $53.81_{\scriptscriptstyle +29.15}$ 
& $51.82_{\scriptscriptstyle +20.91}$ \\

& Qwen2-Audio-7B-Instruct 
& $58.37_{\scriptscriptstyle +4.97}$ 
& $58.63_{\scriptscriptstyle +19.76}$ 
& $53.96_{\scriptscriptstyle +19.80}$ 
& $59.30_{\scriptscriptstyle +14.34}$ 
& $56.66_{\scriptscriptstyle +14.08}$ 
& $60.46_{\scriptscriptstyle +9.57}$ 
& $56.24_{\scriptscriptstyle +21.27}$ 
& $52.02_{\scriptscriptstyle +26.01}$ 
& $52.27_{\scriptscriptstyle +15.91}$ \\

& Audio-Flamingo-3 
& $60.37_{\scriptscriptstyle +2.97}$ 
& $71.25_{\scriptscriptstyle +7.14}$ 
& $64.85_{\scriptscriptstyle +8.91}$ 
& $67.34_{\scriptscriptstyle +6.30}$ 
& $66.00_{\scriptscriptstyle +4.74}$ 
& $63.86_{\scriptscriptstyle +6.17}$ 
& $68.30_{\scriptscriptstyle +9.21}$ 
& $70.85_{\scriptscriptstyle +7.18}$ 
& $59.55_{\scriptscriptstyle +8.63}$ \\

& AF-Next-Think 
& $88.56_{\scriptscriptstyle -25.22}$ 
& $86.66_{\scriptscriptstyle -8.27}$ 
& $83.17_{\scriptscriptstyle -9.41}$ 
& $87.48_{\scriptscriptstyle -13.84}$ 
& $86.68_{\scriptscriptstyle -15.94}$ 
& $85.82_{\scriptscriptstyle -15.79}$ 
& $89.57_{\scriptscriptstyle -12.06}$ 
& $89.24_{\scriptscriptstyle -11.21}$ 
& $85.00_{\scriptscriptstyle -16.82}$ \\

\midrule
\multicolumn{2}{c}{\textbf{Avg. $\Delta$ ($\downarrow$)}} 
& $\mathbf{-2.96}$ 
& $\mathbf{+13.32}$ 
& $\mathbf{+11.76}$ 
& $\mathbf{+7.74}$ 
& $\mathbf{+6.25}$ 
& $\mathbf{+5.11}$ 
& $\mathbf{+11.41}$ 
& $\mathbf{+12.78}$ 
& $\mathbf{+7.16}$ \\

\bottomrule
\end{tabular}
}
\caption{Verification accuracy (\%) across categories, gender, and country. For each \textsc{CoT} result, the subscript reports $\Delta$. Top-4 countries are shown.}
\label{tab:score-combined-delta}
\end{table*}

\noindent \textbf{Fact type.}
The benefit of retrieval and reasoning is consistent across all fact categories, but not uniform. The largest average $Lift$ appears for temporal facts, with a gain of $+11.44$ points, followed by location facts at $+9.51$ points and relation facts at $+6.68$ points. This suggests that year-based claims benefit most from retrieved evidence, likely because the evidence often contains an explicit temporal anchor that can be directly compared against the spoken claim. Relation facts remain the most difficult: they receive the smallest average lift and also have the lowest final accuracy for the strongest model, AF-Next-Think, at $83.17\%$ compared with $88.56\%$ on Year and $86.66\%$ on Location. This indicates that relation verification requires more than retrieving a relevant entity; the model must correctly preserve role direction and compare subject--relation--object structure.

The $\Delta$ table adds an important nuance. Year facts have the smallest remaining gap on average, even becoming negative overall ($-2.96$), because AF-Next-Think strongly surpasses its text-only baseline on this category. In contrast, Location and Relation still show positive average gaps of $+13.32$ and $+11.76$, respectively. Thus, retrieval and reasoning substantially help all fact types, but the text-speech modality gap remains more persistent for semantic categories whose text-only baselines are already high.

\noindent \textbf{Gender.}
\textsc{Transcript-RAG+CoT} yields slightly larger gains on male-subject claims than on female-subject claims. The average $Lift$ over the speech-only baseline is $+10.57$ points for male subjects and $+9.25$ points for female subjects, suggesting that retrieval and reasoning help both groups but improve the male-subject slice somewhat more. The final accuracy is also slightly higher for male subjects, averaging approximately $67.1\%$ compared with $65.4\%$ for female subjects across the four LALMs.

However, this higher absolute performance does not imply that the text-speech modality gap is smaller for male-subject claims. The remaining $\Delta$ to the text-only baseline is actually larger for male subjects ($+7.74$) than for female subjects ($+6.25$). This is because the paired text-only LLM baselines are also stronger on male-subject claims, creating a higher reference point for the speech models to match. Thus, RAG+CoT improves male-subject claims slightly more in absolute terms, but it does not fully close the modality gap for them. Overall, gender-based variation is modest compared with the much larger effects of input modality, retrieval, and reasoning.

\noindent \textbf{Country.}
The country-level results show that \textsc{Transcript-RAG+CoT} improves speech fact verification across all four major country slices, but the degree of gap closure differs. The clearest improvement appears for the USA: it has the largest average $Lift$ over the speech-only baseline ($+9.92$ points) and the smallest remaining gap to the text-only baseline ($\Delta=+5.11$). This indicates that, for US-subject claims, retrieval and reasoning not only improve absolute accuracy but also close the text-speech modality gap most effectively.

England and India show a different pattern. Their final accuracies are high, especially for AF-Next-Think ($89.57\%$ on England and $89.24\%$ on India), but their remaining gaps are larger ($+11.41$ and $+12.78$). This means that the models improve substantially on these slices, but the text-only baselines are also strong, so the speech models still have more ground to recover. In other words, high final accuracy does not necessarily imply that the modality gap is fully closed.

Germany is the weakest slice in absolute terms: it has the lowest average final accuracy and the smallest average $Lift$ ($+8.07$). Its remaining gap ($+7.16$) is smaller than England and India, but this should not be read as better speech verification. Rather, the text-only reference is lower on Germany, making the residual gap smaller. Overall, the country analysis suggests that RAG+CoT is most effective for closing the gap on US-subject claims, while England and India remain challenging relative to their strong text baselines, and Germany receives the least absolute benefit. The dominant trend is still model-driven: standard LALMs remain weak across countries, whereas the thinking-tuned model generalizes much more reliably.

\noindent \textbf{Takeaway.}
The stratified results show that \textsc{Transcript-RAG+CoT} improves speech fact verification broadly, but the gains are not uniform. Fact type is the strongest source of variation: temporal facts benefit most from retrieval and reasoning, while relational facts remain the hardest, likely because they require preserving entity roles and relation direction. Gender effects are comparatively small. Male-subject claims receive slightly larger absolute gains, but this does not translate into a smaller modality gap because text-only baselines are also stronger on this slice. Country-level results show clearer differences in gap closure: US-subject claims benefit most, with both the largest $Lift$ and the smallest remaining $\Delta$, whereas England and India retain larger gaps despite high final accuracy, and Germany receives the least absolute improvement. Overall, the dominant pattern remains model-driven: standard LALMs improve only partially, while the thinking-tuned model generalizes more reliably across factual, gender, and country slices.

\begin{table}[t]
\centering
\small
\resizebox{\columnwidth}{!}{
\begin{tabular}{lcc}
\toprule
\textbf{LALM} &
\textbf{C3 Parseable (\%)} &
\textbf{C5 Parseable (\%)} \\
\midrule
Qwen-Audio-Chat           & 98.8  & 98.8  \\
Qwen2-Audio-7B            & 99.5  & 99.5  \\
Audio-Flamingo-3          & 91.9  & 91.9  \\
Phi-4-multimodal          & 29.8  & 82.4  \\
Audio-Flamingo-next-think & 100.0 & 100.0 \\
\bottomrule
\end{tabular}
}
\caption{Percentage of outputs containing a parseable verdict in the required \texttt{<answer>} field under the CoT conditions C3 and C5. Unparseable outputs are scored as incorrect.}
\label{tab:cot-parseability}
\end{table}

\begin{table}[t]
\centering
\small
\resizebox{\columnwidth}{!}{
\begin{tabular}{llccccccc}
\toprule
\textbf{Cond.} & \textbf{Setting} &
\textbf{Year} & \textbf{Loc.} & \textbf{Rel.} &
\textbf{Avg.} & $\boldsymbol{\Delta}$ \textbf{(\(\downarrow\))} &
\textbf{Lift (\(\uparrow\))} &
$\boldsymbol{\Delta_{\mathrm{UB}}}$ \textbf{(\(\downarrow\))} \\
\midrule
C0 & $\mathbf{T}_{\mathrm{LLM}}$
   & 63.34 & 78.39 & 73.76 & 71.8 & --    & --   & --   \\
UB & $\mathbf{T}_{\mathrm{LLM}}^{r_{\mathrm{text}}}$
   & 92.63 & 93.62 & 91.09 & 92.4 & --    & --   & --   \\
\midrule
C1 & $\mathbf{T}_{\mathrm{LALM}}$
   & 60.79 & 73.18 & 65.35 & 66.4 & 5.4   & --   & 26.0 \\
C2 & $\mathbf{S}_{\mathrm{LALM}}$
   & 63.06 & 69.72 & 66.34 & 66.4 & 5.4   & --   & 26.0 \\
C3 & $\mathbf{S}_{\mathrm{LALM}}[\mathrm{CoT}]$
   & 68.44 & 74.62 & 72.28 & 71.8 & 0.0   & 5.4  & 20.6 \\
C4 & $\mathbf{S}_{\mathrm{LALM}}^{r_{\mathrm{text}}}$
   & 82.43 & 81.63 & 80.20 & 81.4 & -9.6  & 15.0 & 11.0 \\
C5 & $\mathbf{S}_{\mathrm{LALM}}^{r_{\mathrm{text}}}[\mathrm{CoT}]$
   & 88.56 & 86.66 & 83.17 & 86.1 & -14.3 & 19.7 & 6.3  \\
\bottomrule
\end{tabular}}
\caption{Category-wise accuracy (\%) of Audio-Flamingo-next-think across all settings.}
\label{tab:af-next-all-settings}
\end{table}

\begin{table*}[t]
\centering
\footnotesize
\setlength{\tabcolsep}{4pt}
\renewcommand{\arraystretch}{1.15}
\begin{tabularx}{\textwidth}{@{}l l l X c c c@{}}
\toprule
Model & Category & Subject & Claim & GT & $\mathbf{S}_\mathrm{LALM}$ & $\mathbf{S}_\mathrm{LALM}^{\mathbf{r}_\mathrm{text}}$ \\
\midrule
\multicolumn{7}{@{}l}{\textit{RAG helps: retrieval flips a wrong no-RAG answer to correct.}}\\
\midrule
AF-3 & Year & Magdalena Neuner & Magdalena Neuner married in 2014. & Yes & No & Yes \\
AF-3 & Location & Mauricio Isla & Mauricio Isla is from Chile. & Yes & No & Yes \\
AF-3 & Relation & Jeb Bush & Barbara is Jeb Bush's mother. & Yes & No & Yes \\
\midrule
\multicolumn{7}{@{}l}{\textit{RAG hurts: retrieval flips a correct no-RAG answer to wrong.}}\\
\midrule
AF-3 & Year & Kesha & Kesha was 18 years old in 2005. & Yes & Yes & No \\
AF-3 & Location & Lionel Messi & Lionel Messi is from Argentina. & Yes & Yes & No \\
AF-3 & Relation & Derek Trucks & Derek Trucks is married to Susan Tedeschi. & Yes & Yes & No \\
\bottomrule
\end{tabularx}
\caption{A selection of specific failure cases where retrieval flips prediction.}
\label{tab:rag_impact_examples}
\end{table*}

\begin{table*}[t]
\centering
\footnotesize
\setlength{\tabcolsep}{4pt}
\renewcommand{\arraystretch}{1.25}
\begin{tabularx}{\textwidth}{@{}l X@{}}
\toprule
\textbf{Setting} & \textbf{Prompt template} \\
\midrule

\multicolumn{2}{@{}l}{\textit{Audio LALM input (speech is the claim)}}\\
\addlinespace[1pt]

\hdashline

$\mathbf{S}_{\mathrm{LALM}}$ (C2)
& \texttt{Is the content in the speech factually correct? Respond with only Yes or No.} \\

\addlinespace[2pt]
$\mathbf{S}^{\mathbf{r}_{\mathrm{text}}}_{\mathrm{LALM}}$ (C4)
& \texttt{Given the additional context: \ph{knowledge}. Is the content in the speech factually correct? Respond with only Yes or No.} \\

\addlinespace[2pt]
$\mathbf{S}_{\mathrm{LALM}}[\textsc{CoT}]$ (C3)
& \texttt{Listen to the audio carefully. First, transcribe the spoken claim verbatim inside <claim>...</claim> tags. Then reason step by step about whether the claim is factually correct. Finally, write only the word "Yes" or "No" inside <answer>...</answer> tags.} \\

\addlinespace[2pt]
$\mathbf{S}^{\mathbf{r}_{\mathrm{text}}}_{\mathrm{LALM}}[\textsc{CoT}]$ (C5)
& \texttt{You are given the following retrieved background information: \ph{knowledge}. Listen to the audio carefully. First, transcribe the spoken claim verbatim inside <claim>...</claim> tags. Then reason step by step about whether the claim is factually correct, using both the audio and the background information above. Finally, write only the word "Yes" or "No" inside <answer>...</answer> tags.} \\

\midrule

\multicolumn{2}{@{}l}{\textit{Text LLM input (claim is the transcript)}}\\
\addlinespace[1pt]
\hdashline

$\mathbf{T}_{\mathrm{LLM}}$ (C0)
& \texttt{Is the following text factually correct? Respond with only 'Yes' or 'No'.\textbackslash nText: \ph{transcript}} \\

\addlinespace[2pt]
$\mathbf{T}^{\mathbf{r}_{\mathrm{text}}}_{\mathrm{LLM}}$
& \texttt{Given the additional context: \ph{knowledge}. Is the following text factually correct? Respond with only 'Yes' or 'No'.\textbackslash nText: \ph{transcript}} \\

\midrule

\multicolumn{2}{@{}l}{\textit{Audio LALM run in text-only mode (claim fed as text, audio dropped)}}\\
\addlinespace[1pt]
\hdashline

$\mathbf{T}_{\mathrm{LALM}}$ (C1)
& \texttt{Is the following text factually correct? Respond with only Yes or No.\textbackslash nText: \ph{transcript}} \\

\bottomrule
\end{tabularx}
\caption{Prompt templates used across the experimental settings in VeriSpeak. \ph{knowledge} is the top-$k$ retrieved evidence concatenated from the retriever output. \ph{transcript} is either the oracle written claim or its ASR output, depending on the evaluation condition. The \emph{CoT} prompts elicit a \texttt{<claim>}/\texttt{<answer>}-tagged reasoning trace, while the non-CoT prompts elicit a direct binary verdict.}
\label{tab:prompts}
\end{table*}

\section{Additional Failure Modes}
\label{app:output-diagnostics}

We conduct output-level diagnostics to better understand the near- and below-random performance observed in several model–setting pairs.

\paragraph{Constant-answer behavior.}
We inspect the predicted verdicts for every model--setting pair. We observe one complete collapse: Qwen2-Audio-7B under the speech-only condition C2 predicts \textit{No} for every sample, resulting in 49.2\% accuracy. Other model--setting pairs with accuracy near 50\% produce both verdict classes and therefore cannot be explained by constant-answer behavior alone.

\paragraph{Output parseability.}
Our CoT prompts require the final verdict to appear as \textit{Yes} or \textit{No} inside an \texttt{<answer>} field. We apply strict parsing: if this field is missing or does not contain a parseable verdict, the output is marked incorrect. Table~\ref{tab:cot-parseability} reports the percentage of outputs satisfying this requirement under the speech-only CoT condition C3 and the \textsc{Transcript-RAG+CoT} condition C5.

Most model--setting pairs have parseability rates above 90\%. The main exception is Phi-4-multimodal under C3, for which only 29.8\% of outputs are parseable. Inspection shows that the model frequently terminates after generating the \texttt{<claim>} field without producing the required \texttt{<answer>} field. Its below-random C3 accuracy therefore partly reflects format non-compliance and should not be interpreted solely as factual-verification performance. Providing retrieved evidence in C5 raises its parseability to 82.4\%.

For the Qwen models, parseability remains approximately 99\%, yet C3 does not improve over the corresponding non-CoT speech-only condition C2. Audio-Flamingo-3 exhibits a similar pattern despite a parseability rate above 90\%. Thus, the weak C3 results cannot generally be attributed to formatting failures: vanilla CoT remains unreliable for recovering factual-verification ability from speech, while Phi-4-multimodal additionally exhibits a model-specific format-following failure.

\begin{table*}[t]
\centering
\footnotesize

\resizebox{\textwidth}{!}{%
\begin{tabular}{lllccccc}
\toprule
Cat. & Subject & Claim & GT
& $\mathbf{S}_{\mathrm{LALM}}$
& $\mathbf{S}_{\mathrm{LALM}}[\mathrm{CoT}]$
& $\mathbf{S}_{\mathrm{LALM}}^{\mathbf{r}_{\mathrm{text}}}$
& $\mathbf{S}_{\mathrm{LALM}}^{\mathbf{r}_{\mathrm{text}}}[\mathrm{CoT}]$ \\
\midrule

Year & Jordin Sparks
& Jordin Sparks rose to fame in 2011. \newline {\footnotesize (Original fact: Jordin Sparks rose to fame in 2007.)}
& No & $\xmark$ & $\xmark$ & $\xmark$ & $\cmark$ \\

\addlinespace

Loc. & John Lydon
& John Lydon is from Belfast. \newline {\footnotesize (Original fact: John Lydon is from the United Kingdom.)}
& No & $\xmark$ & $\xmark$ & $\xmark$ & $\cmark$ \\

\addlinespace

Rel. & Susan Downey
& Susan Downey is co-president of Belfast. \newline {\footnotesize (Original fact: Susan Downey is co-president of Dark Castle Entertainment.)}
& No & $\xmark$ & $\xmark$ & $\xmark$ & $\cmark$ \\

\bottomrule
\end{tabular}
}

\caption{Qualitative cases where retrieval combined with CoT is the only configuration that fixes Audio-Flamingo-3. Each row shows a claim where the speech-only setting is wrong, CoT alone is wrong, transcript-based RAG alone is wrong, but retrieval with CoT recovers the correct verdict. \textbf{GT}: \textit{No} = factually incorrect.}
\label{tab:qual-af3-progression}
\end{table*}

\begin{table*}[t]
\centering
\footnotesize
\setlength{\tabcolsep}{5pt}
\renewcommand{\arraystretch}{1.20}

\resizebox{\textwidth}{!}{%
\begin{tabular}{@{}ll>{\raggedright\arraybackslash}p{9.5cm}ccccc@{}}
\toprule
Cat. & Subject & Claim & GT
& $\mathbf{S}_{\mathrm{LALM}}$
& $\mathbf{S}_{\mathrm{LALM}}[\mathrm{CoT}]$
& $\mathbf{S}_{\mathrm{LALM}}^{\mathbf{r}_{\mathrm{text}}}$
& $\mathbf{S}_{\mathrm{LALM}}^{\mathbf{r}_{\mathrm{text}}}[\mathrm{CoT}]$ \\
\midrule

Year & Jillian Michaels
& Jillian Michaels hosted in fall 2015.
& Yes & $\xmark$ & $\xmark$ & $\xmark$ & $\cmark$ \\

\addlinespace[0.35em]

Year & Audrina Patridge
& Audrina Patridge rose to prominence in 2013. \newline
{\footnotesize (Original fact: Audrina Patridge rose to prominence in 2006.)}
& No & $\xmark$ & $\xmark$ & $\xmark$ & $\cmark$ \\

\addlinespace[0.35em]

Loc. & Ciara
& Ciara is of Irish origin.
& Yes & $\xmark$ & $\xmark$ & $\xmark$ & $\cmark$ \\

\addlinespace[0.35em]

Loc. & J-Ax
& J-Ax is a South Korean singer. \newline
{\footnotesize (Original fact: J-Ax is an Italian singer.)}
& No & $\xmark$ & $\xmark$ & $\xmark$ & $\cmark$ \\

\addlinespace[0.35em]

Rel. & Kelly Ripa
& Kelly Ripa is married to Mark Consuelos.
& Yes & $\xmark$ & $\xmark$ & $\xmark$ & $\cmark$ \\

\addlinespace[0.35em]

Rel. & Camila Alves
& Camila Alves is married to Wynton Marsalis. \newline
{\footnotesize (Original fact: Camila Alves is married to Matthew McConaughey.)}
& No & $\xmark$ & $\xmark$ & $\xmark$ & $\cmark$ \\

\bottomrule
\end{tabular}%
}

\caption{Qualitative cases where only retrieval combined with CoT predicts the label correctly on AF-next-think. Original fact is given wherever GT is No.}
\label{tab:qual-afnt-progression}
\end{table*}

\begin{table*}[t]
\centering
\footnotesize
\setlength{\tabcolsep}{5pt}
\renewcommand{\arraystretch}{1.25}

\resizebox{\textwidth}{!}{%
\begin{tabular}{@{}ll>{\raggedright\arraybackslash}p{10cm}cccc@{}}
\toprule
Category & Subject & Claim & GT
& $\mathbf{S}_{\mathrm{AF\text{-}3}}$
& $\mathbf{S}_{\mathrm{AF\text{-}3}}^{\mathbf{r}_{\mathrm{text}}}$
& $\mathbf{S}_{\mathrm{AF\text{-}NT}}$ \\
\midrule

Year
& Sarah Palin
& Sarah Palin was born in 1974. \newline
{\footnotesize (Original fact: Sarah Palin was born in 1964.)}
& No & Yes & Yes & No \\

\addlinespace[0.45em]

Year
& Bar Refaeli
& Bar Refaeli was born in 1988. \newline
{\footnotesize (Original fact: Bar Refaeli was born in 1985.)}
& No & Yes & Yes & No \\

\addlinespace[0.45em]

Loc.
& Yuto Nagatomo
& FC Tokyo is a Japanese club.
& Yes & No & No & Yes \\

\addlinespace[0.45em]

Loc.
& Edinson Cavani
& Edinson Cavani is from Oman. \newline
{\footnotesize (Original fact: Edinson Cavani is from Uruguay.)}
& No & Yes & Yes & No \\

\addlinespace[0.45em]

Rel.
& Sofia Coppola
& Sofia Coppola is the sister of Francis Ford Coppola. \newline
{\footnotesize (Original fact: Sofia Coppola is the daughter of Francis Ford Coppola.)}
& No & Yes & Yes & No \\

\addlinespace[0.45em]

Rel.
& Lily Allen
& Lily Allen is the daughter of Namitha. \newline
{\footnotesize (Original fact: Lily Allen is the daughter of Keith Allen.)}
& No & Yes & Yes & No \\

\bottomrule
\end{tabular}%
}

\caption{Qualitative examples where AF-3 and AF-3 with \textsc{transcript-RAG} fail, while AF-Next-Think succeeds without RAG. Original fact is given wherever GT is No.}
\label{tab:qualitative_examples}
\end{table*}

\section{Detailed Results for AF-next-think}
\label{app:af-next-results}

For completeness, we report category-wise results for Audio-Flamingo-next-think across all evaluation settings in Table~\ref{tab:af-next-all-settings}. 

\section{Atomic Fact Extraction Prompts}
\label{app:atomic-prompts}

Section~\ref{sec:verispeak_dataset} extracts one atomic, pronoun-free factual
sentence per claim from each flagged bio using
Llama-3.2-3B-Instruct\footnote{\texttt{meta-llama/Llama-3.2-3B-Instruct}}(
\texttt{max\_new\_tokens=512}, \texttt{do\_sample=True},
\texttt{temperature=0.1}). All three categories share a common
system message and differ only in the user prompt. Substitution
variables are pulled per row from the enriched knowledge base:
\texttt{\{name\}} is the celebrity name, \texttt{\{text\}} is the bio,
and \texttt{\{evidence\}} is the matched signal from the upstream
flagging stage (regex year matches, spaCy GPE/LOC spans, or regex
relation keywords). Outputs are post-filtered to remove facts containing fewer than four
words and facts containing unresolved pronouns from
\{\textit{he, she, his, her, him, they, it, their}\}, which are not
self-contained. We retain locally coreferential possessives when the
antecedent is explicitly named in the same sentence (e.g., ``Nelly
embarked on his music career with Midwest hip hop group St. Lunatics
in 1993.'').

\paragraph{Shared system message.}
\begin{quote}\ttfamily\small
You are a professional fact extraction bot that only outputs atomic,
pronoun-free sentences based on the provided text.
\end{quote}

\paragraph{Year prompt.}
\begin{quote}\ttfamily\small
You are a fact extraction assistant. Given the text below about
\{name\}, extract only the atomic factual sentences that describe a
specific event involving a year or time period (like \{evidence\}).\\[2pt]
Text: ``\{text\}''\\[2pt]
Rules:\\
1. Each sentence must be atomic (containing exactly one fact).\\
2. Each sentence MUST contain at least one year or time period mentioned in the text.\\
3. Replace all pronouns (he, she, they, his, her, their, him) with `\{name\}'.\\
4. Do not use any knowledge outside of the provided text.\\
5. Do not number the sentences or add headings.\\
6. If no year-based facts are found, return an empty list.\\[2pt]
Example:\\
Text: ``Herman Van Rompuy served as Prime Minister of Belgium from
2008 to 2009. He later became President of the European Council in
2009.''\\
Output:\\
Herman Van Rompuy served as Prime Minister of Belgium from 2008 to 2009.\\
Herman Van Rompuy became President of the European Council in 2009.\\[2pt]
Output for \{name\}:
\end{quote}

\paragraph{Location prompt.}
\begin{quote}\ttfamily\small
You are a fact extraction assistant. Given the text below about
\{name\}, extract only the atomic factual sentences that describe a
specific location (city, country, school, landmark, etc.\ like
\{evidence\}) where \{name\} lived, worked, studied, or achieved
something.\\[2pt]
Text: ``\{text\}''\\[2pt]
Rules:\\
1. Each sentence must be atomic (containing exactly one fact).\\
2. Each sentence MUST contain a specific location mentioned in the text.\\
3. Replace all pronouns (he, she, they, his, her, their, him) with `\{name\}'.\\
4. Do not use any knowledge outside of the provided text.\\
5. Do not number the sentences or add headings.\\
6. If no location facts are found, return an empty list.\\[2pt]
Example:\\
Text: ``Sundar Pichai was born in Madras, India. He later moved to
the United States for studies at Stanford.''\\
Output:\\
Sundar Pichai was born in Madras, India.\\
Sundar Pichai moved to the United States.\\
Sundar Pichai studied at Stanford.\\[2pt]
Output for \{name\}:
\end{quote}

\paragraph{Relation prompt.}
\begin{quote}\ttfamily\small
You are a fact extraction assistant. Given the text below about
\{name\}, extract only the atomic factual sentences that describe a
specific relation (family member, spouse, partner, child, parent,
sibling, colleague, or professional associate like \{evidence\})
between \{name\} and another person or entity.\\[2pt]
Text: ``\{text\}''\\[2pt]
Rules:\\
1. Each sentence must be atomic (containing exactly one fact about a relation).\\
2. Each sentence MUST contain a specific person or entity that \{name\} has a relation with.\\
3. Replace all pronouns (he, she, they, his, her, their, him) with `\{name\}'.\\
4. Do not use any knowledge outside of the provided text.\\
5. Do not number the sentences or add headings.\\
6. If no relation facts are found, return an empty list.\\[2pt]
Example:\\
Text: ``Barack Obama is married to Michelle Obama. They have two
daughters, Malia and Sasha. He worked closely with Joe Biden during
his presidency.''\\
Output:\\
Barack Obama is married to Michelle Obama.\\
Barack Obama has a daughter named Malia.\\
Barack Obama has a daughter named Sasha.\\
Barack Obama worked closely with Joe Biden.\\[2pt]
Output for \{name\}:
\end{quote}

\paragraph{Negative-fact generation.}
The ``incorrect'' counterparts are produced by deterministic
perturbation rather than prompting, ensuring tight contrast with each
correct sentence:
\begin{itemize}
  \item \textbf{Year.} Each 4-digit year matched by the regex
    \texttt{\textbackslash b(1\textbackslash d\{3\}|20[0-2]\textbackslash d)\textbackslash b}
    is shifted by a non-zero offset drawn uniformly from
    $[-10, +10]$, clamped to remain a plausible 4-digit year.
  \item \textbf{Location.} spaCy \texttt{en\_core\_web\_sm} extracts
    \texttt{GPE}/\texttt{LOC}/\texttt{NORP} spans and swaps each one
    with a same-category candidate drawn from a typed pool built
    from the KB (countries from the enriched \texttt{country} field;
    cities from \texttt{location evidence}; a fixed nationality list
    for \texttt{NORP}). Spans overlapping the celebrity's name are
    skipped.
  \item \textbf{Relation.} A coin flip selects between
    (i)~replacing a relation keyword via a hand-written map
    (e.g.\ \texttt{married to} $\rightarrow$ \texttt{divorced from},
    \texttt{son of} $\rightarrow$ \texttt{father of},
    \texttt{attended} $\rightarrow$ \texttt{dropped out of},
    \texttt{born in} $\rightarrow$ \texttt{died in}), and
    (ii)~swapping a spaCy \texttt{PERSON}/\texttt{ORG} span with a
    same-type candidate from KB-derived pools, never the celebrity
    themselves.
\end{itemize}
Sentences for which no negative could be produced are dropped so that
the per-celebrity correct/incorrect counts stay balanced.

\section{Additional Qualitative Samples (Visual)}

Tables~\ref{tab:qual-af3-progression},~\ref{tab:qual-afnt-progression},~\ref{tab:qualitative_examples} present visual qualitative examples illustrating model behavior across the evaluated settings.

\section{Implementation Details}
\label{sec:appendix}

We ran all experiments using PyTorch and the Hugging Face Transformers library. For most LLMs and LALMs considered in this work, we used either the authors' original code repositories or their Hugging Face implementations, depending on availability and ease of reproducibility. The models and their parameter sizes are summarized in Table~\ref{tab:model_params}. All experiments were conducted on a machine equipped with three NVIDIA A6000 GPUs, each with 48 GB of memory. The reported results are averaged over three runs.

\section{AI Use Statement}

We used AI assistance only for polishing the manuscript writing and improving the visual presentation of Figure~\ref{fig:teaser_fig}. All ideas, experiments, analyses, and conclusions are our own.

\begin{table}[t]
\centering
\small
\setlength{\tabcolsep}{4pt}
\renewcommand{\arraystretch}{1.15}
\resizebox{\columnwidth}{!}{%
\begin{tabular}{@{}llr@{}}
\toprule
\textbf{Family} & \textbf{Model} & \textbf{\#Params} \\
\midrule
\multirow{4}{*}{Qwen} 
& Qwen-7B (LLM backbone)        & 7B \\
& Qwen-Audio-Chat               & 7-8B \\
& Qwen2-Audio-7B-Instruct       & 8.2B \\
\midrule
\multirow{3}{*}{Qwen2.5} 
& Qwen2.5-7B-Instruct (LLM backbone) & 7.6B \\
& Audio-Flamingo-3              & 7-8B \\
& Audio-Flamingo-next-think     & 8B \\
\midrule
\multirow{2}{*}{Phi} 
& Phi-4-mini-instruct (LLM backbone) & 3.8B \\
& Phi-4-multimodal-instruct     & 5.6B \\
\bottomrule
\end{tabular}%
}
\caption{Models evaluated in our experiments with approximate parameter counts.}
\label{tab:model_params}
\end{table}

\end{document}